\documentclass[11pt]{article}
\usepackage{natbib}

\usepackage{titlesec}

\usepackage{graphicx}

\usepackage{amsthm}
\usepackage{amssymb}  
\usepackage{amsmath} 
\usepackage[font={small}]{caption}
\usepackage{subcaption}
\usepackage{url}
\usepackage{algorithmic}
\usepackage{algorithm}
\usepackage{xcolor}
\usepackage{eqparbox}
\usepackage{xspace}
\usepackage{lastpage}
\usepackage{authblk}
\usepackage{fullpage}
\usepackage{footnote}

\floatname{algorithm}{}

\newtheorem{remark}{Remark}

\newtheorem{problem}{Problem}

\newcommand{\x}{\mathcal X}
\newcommand{\xfree}{\mathcal X_{\text{free}}}
\newcommand{\xobs}{\mathcal X_{\text{obs}}}
\newcommand{\xgoal}{\mathcal X_{\text{goal}}}
\newcommand{\xinit}{x_{\text{init}}}

\newcommand{\prm}{PRM$^\ast$\xspace}
\newcommand{\rrt}{RRT$^\ast$\xspace}
\newcommand{\fmt}{FMT$^\ast$\xspace}
\newcommand{\gmt}{GMT$^\ast$\xspace}
\newcommand{\bit}{BIT$^\ast$\xspace}

\graphicspath{{./fig/}}

\title{Learning Sampling Distributions for Robot Motion Planning}

\author[1,\thanks{
These authors contributed equally to this work. 
\\ \indent ichter@google.com, jharrison@stanford.edu, pavone@stanford.edu
\\ \\ \indent This work was originally presented at the 2018 IEEE International Conference on Robotics and Automation (ICRA). This extended version includes new numerical experiments demonstrating generalization, iterative retraining of the generative model, and multirobot planning experiments.
}]
{Brian Ichter}
\author[2,$^*$]{James Harrison}
\author[3]{Marco Pavone}
\affil[1]{Google Brain}
\affil[2]{Department of Mechanical Engineering, Stanford University}
\affil[3]{Department of Aeronautics and Astronautics, Stanford University}
\date{}
\begin{document}

\maketitle


\begin{abstract}
A defining feature of sampling-based motion planning is the reliance on an implicit representation of the state space, which is enabled by a set of probing samples. Traditionally, these samples are drawn either probabilistically or deterministically to {\em uniformly} cover the state space. Yet, the motion of many robotic systems is often restricted to ``small" regions of the state space, due to, for example, differential constraints or collision-avoidance constraints. To accelerate the planning process, it is thus desirable to devise {\em non-uniform} sampling strategies that favor sampling in those regions where an optimal solution might lie. This paper proposes a methodology for non-uniform sampling, whereby a sampling distribution is {\em learned} from demonstrations, and then used to bias sampling.  
The sampling distribution is computed through a conditional variational autoencoder, allowing sample generation from the latent space conditioned on the specific planning problem.  
This methodology is general, can be used in combination with any sampling-based planner, and can effectively exploit the underlying structure of a planning problem while maintaining the theoretical guarantees of sampling-based approaches. Specifically, on several planning problems, the proposed methodology is shown to effectively learn representations for the relevant regions of the state space, resulting in an order of magnitude improvement in terms of success rate and convergence to the optimal cost.
\end{abstract}


\section{Introduction}

Sampling-based motion planning (SBMP) has emerged as a successful algorithmic paradigm for solving high-dimensional, complex, and dynamically-constrained motion planning problems. A defining feature of SBMP is the reliance on an implicit representation of the state space, achieved through sampling the feasible space and probing local connections through a black-box collision checking module.  Traditionally, these samples are drawn either probabilistically or deterministically to {\em uniformly} cover the state space.
Such a sampling approach allows arbitrarily accurate representations (in the limit of the number of samples approaching infinity), and thus allows theoretical guarantees on completeness and asymptotic optimality. In practice, many robotic systems only operate in small subsets of the state space, which may be very complex and only implicitly defined. This might be due to the  environment (e.g., initial and goal conditions, narrow passageways), the system's dynamics (e.g., a bipedal walking robot's preference towards stable, upright conditions), or implicit constraints (e.g., loop closures, multi-robot systems remaining out of collision). 
The performance of SBMP is thus tied to the placement of samples in these promising regions, a result uniform sampling is only able to achieve through sheer exhaustion. 
Therein lies a fundamental challenge of SBMP algorithms: while their implicit representation of the state space is very general and requires only minimal assumptions, it limits their ability to leverage knowledge gained from previous planning problems or to use known information about the workspace or robotic system to accelerate solutions. 

\begin{figure}[htb]
\centering
\includegraphics[width=0.5\textwidth]{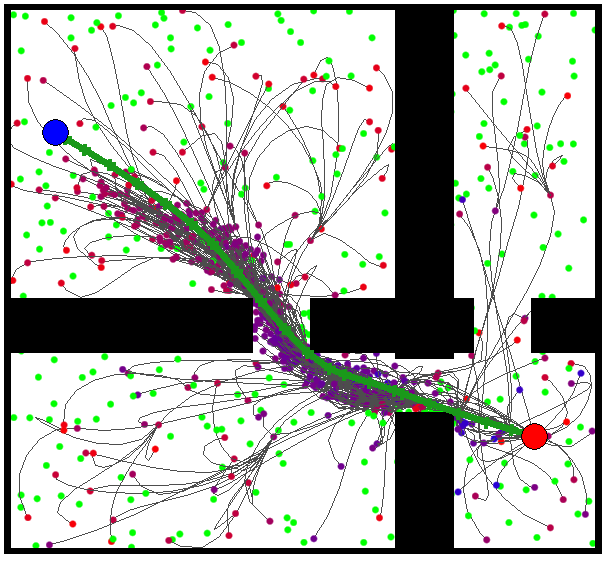}
\caption{A fast marching tree (\fmt) generated with learned samples for a double integrator, conditioned on the initial state (red circle), goal region (blue circle), and workspace obstacles (black). Note the significantly higher density of samples in the region around the solution.}
\label{fig:learnedTreeNarrow}
\end{figure}

In this work we approach this challenge through biasing the sampling of the state space towards these promising regions via {\em learned} sample distributions (see Fig. \ref{fig:learnedTreeNarrow}). 
At the core of this methodology is a conditional variational autoencoder (CVAE) \citep{sohn2015learning}, which is capable of learning complex manifolds and the regions around them, and is trained from demonstrations of successful motion plans and previous robot experience.
The latent space of the CVAE can then be sampled and projected into a representation of the promising regions of the state space, conditioned on information specific to a given planning problem, for example, initial state, goal region, and obstacles (as defined in the workspace). 
In this way one may maintain the theoretical and exploration benefits of sampling-based motion planning, while leveraging the generality, extensibility, and practical performance associated with hyperparametric learning approaches. 
Remarkably, this methodology is extensible to virtually any system and available problem-specific information.

\paragraph{Statement of Contributions.} The contributions of this paper are threefold. 
First, we present a methodology to learn sampling distributions for sampling-based motion planning. 
Such learned sampling distributions can be used to focus sampling on those regions where an optimal solution may lie, as dictated by both system-specific and problem-specific constraints. 
This methodology is general to arbitrary systems and problems, scales well to high dimensions, and maintains the typical theoretical guarantees of SBMP (chiefly, completeness and asymptotic optimality). 
Second, we demonstrate the learned sampling distribution methodology on a wide variety of problems, from low-dimensional to high-dimensional, and from a geometric setting to problems with complex dynamics.  
Third, we compare our methodology to uniform sampling, finding approximately an order of magnitude improvement in success rates and path costs, as well as to state of the art approaches for sample biasing. 
Our findings show analogous benefits to those seen recently in the computer vision community, where feature learning-based approaches have resulted in substantially improved performance over human intuition and handcrafted features.

\paragraph{Organization.} This paper is organized as follows.
The next section reviews related work in non-uniform and learned measures for sampling.
Section \ref{sec:problemstatement} reviews the problem addressed herein.
Section \ref{sec:method} outlines the learned sampling distribution methodology.
Section \ref{sec:numexp} demonstrates the performance of the method through numerical experiments on a variety of environments. 
Section \ref{sec:gen} experimentally investigates the method's hyperparameters, potential data sources, generalization, and extensions.
Section \ref{sec:conclusions} summarizes the results and provides directions for future work.

\section{Related Work}

In this section, we review approaches to improving the efficiency of sampling-based motion planning via non-uniform sampling schemes. We divide these methods into four groups: heuristic methods, informed sampling, adaptive sampling, and learning-based approaches. First, we review heuristics for biased sampling, which use, for example, workspace information to choose sampling measures. Informed sampling refers to techniques which alter the sampling distribution based on the current best trajectory, whereas adaptive sampling alters the distribution based on the previously taken samples. These methods are both online methods (referring to the distribution changing online), whereas the heuristic methods are primarily offline, and thus the distribution does not change during the motion planning problem. An important distinction between these two approaches is that informed sampling is not able to refine the sampling distribution until a feasible trajectory has been found. Finally, we review learning methods for sampling, which spans both online and offline methods. The approach presented in this paper is an offline, learning-based method. 

\paragraph{Heuristics for Biased Sampling.} Several previous works have presented non-uniform sampling strategies for SBMP, often resulting in significant performance gains \citep{urmson2003approaches, hsu2006probabilistic}. Other works leverage models of the workspace to bias samples via decomposition techniques \citep{kurniawati2004workspace,van2005using} or via approximation of the medial axis \citep{yang2004adapting}.
While these sampling heuristics are effective when used within the problems for which they were designed, it is often unclear how they perform on environments that are outside of their expected operating conditions \citep{geraerts2004sampling}.
This is particularly the case for planning problems that must sample the full state space (e.g., that include velocity).
These methods further require significant expert intuition, while our method is able to compute efficient non-uniform distributions from demonstration only.
Indeed, this class of biased sampling algorithms does not learn from previous sampling problems or adapt during a problem. 

\paragraph{Informed Sampling.} Recent work by \citet{gammell2014informed} places batches of samples based on information gained from the running best solution during the motion planning problem. 
This approach was generalized in \citep{gammell2015batch} and \citep{choudhury2016regionally}, but these approaches are restricted to geometric motion planning problems. 
These approaches, which reject samples if they could not improve the current best solution, are herein referred to as informed sampling approaches \citep{gammell2014informed,karaman2011anytime}.
In the case of geometric motion planning problems with Euclidean cost, informed sampling methods generate an elliptical informed set which may be sampled from directly \citep{ferguson2006anytime}.
For problems with differential constraints, steering solutions must be computed and thus a closed-form representation of the sampling region does not exist. 
Samples may be rejected if they are guaranteed to not improve the current trajectory \citep{akgun2011sampling}, but this leads to a large fraction of samples being rejected in high dimensional problems \citep{kunz2016hierarchical}.
\citet{kunz2016hierarchical} and \citet{yi2017generalizing} improve the efficiency of this approach using hierarchical rejection sampling and Markov Chain Monte Carlo methods. 
Generally, informed sampling approaches aim to leverage the current best path to improve sampling quality. However, this requires a solution to the motion planning problem before the sampling may be refined.

\paragraph{Adaptive Sampling.} Adaptive sampling typically refers to techniques that alter the sampling distribution based on knowledge of the environment gained from previous samples, during the same motion planning problem.
Several approaches to adaptive sampling have used online learning to improve generalization. 
\citet{burns2005toward} aim to optimally sample the configuration space based on maximizing a given utility function for each new sample, related to the quality of configuration space representation. 
\citet{coleman2015experience} use a related experience-based approach to store experiences in a graph rather than as individual paths.
\citet{kumar2010adaptive} leverage information encoded in local connectivity graphs to improve sampling efficiency. 
\citet{burns2005sampling} aim to address the narrow corridor problem by classifying points in the configuration space as either free space or obstructed space, and the latter are sampled more densely.
Each of these approaches either fails to generalize to arbitrary planning problems or cannot leverage both the full state (e.g., velocity information) and external (e.g., obstacle information) factors. 
In this work we propose the use of recent advances in representation learning to learn sampling distributions. These models are able to include all available state and problem information due to their ability to represent general, high-dimensional, complex conditional distributions.

\paragraph{Learned Sampling.} Finally, we use the term \textit{learning} to refer to the case where knowledge of previous planning problems is leveraged to improve the quality of samples in a given motion planning problem. This is in contrast to informed sampling, which only improves efficiency once a solution is found, and adaptive sampling, which only adjusts sampling distributions after information has been gained from previous samples. Note that informed, adaptive, and learned sampling are sometimes used interchangeably, but we will consistently use these terms as defined above in this work. Learning approaches have the potential to dramatically improve the rate at which good solutions may be found, as they do not require information to be acquired in a given motion planning problem before refining sample placement. 
\citet{zucker2008adaptive} use a reinforcement learning approach to learn where to sample in a discretized workspace. 
This approach is based on a discretization of the workspace. 
While discretizing the workspaces avoids the severe impact of the curse of dimensionality associated with a complex configuration space, it may substantially degrade sampling performance versus learning a sampling distribution in the state space. 
\citet{berenson2012robot} cache paths, and use a learned heuristic to store new paths as well as recall and repair old paths. 
\citet{li2011learning} learn cost-to-go metrics for kinodynamic systems, which often do not have an inexpensive-to-compute metric, especially for nonholonomic systems. 
\citet{baldwin2010non} learn distributions that leverage semantic information. 
\citet{ye2017guided} present a learning-based approach that leverages human demonstrations and SBMP to generate plans.
Recently, \citet{lehner2017repetition} fit Gaussian Mixture Models to previous solution configurations. This model, however, is restrictive and may not generalize as well as the CVAE architecture used herein. 

\section{Problem Statement}\label{sec:problemstatement}

The goal of this work is to generate sample distributions to aid sampling-based motion planning algorithms in solving the optimal motion planning problem.
Informally, solving this problem entails finding the lowest-cost free trajectory from an initial state to a goal region, if one exists.
The simplest version of the problem, referred to herein as the geometric motion planning problem, is defined as follows.
Let $\x = [0,1]^d$ be the state space, with $d\in \mathbb{N},d\geq 2$.
Let $\xobs$ denote the obstacle space, $\xfree = \x \setminus \xobs$ the free state space, $\xinit\in\xfree$ the initial condition, and $\xgoal \subset \xfree$ the goal region.
A path is defined by a continuous function, $s: [0,1] \to \mathbb{R}^d$.
We refer to a path as \emph{collision-free} if $s(\tau) \in \xfree$ for all $\tau \in [0,1]$, and \emph{feasible} if it is collision-free, $s(0) = \xinit$, and $s(1) \in \xgoal$. 
For the geometric motion planning problem, we consider the cost $c$ as the Euclidean distance.
We thus wish to solve, 

\vspace{0.15cm}
\begin{problem}[Optimal motion planning]
Given a motion planning problem $(\xfree, \xinit, \xgoal)$ and a cost $c$, find a feasible path $s^\ast$ such that $c(s^\ast) = \min \{c(s):s $ is feasible$\}$. If no such path exists, report failure.
\end{problem}
\vspace{0.15cm}

Generally, there exist formulations of this problem for systems with kinematic, differential, or more complex constraints, for which we refer the reader to \citep{SchmerlingJansonEtAl2015,lavalle2006planning}. The general form of the motion planning problem is known to be PSPACE-complete \citep{lavalle2006planning}, and thus one often turns to approximate methods to solve the problem efficiently.
Sampling-based motion planning has achieved particular success in solving complex, high-dimensional planning problems.
These algorithms (e.g., \prm, \rrt \citep{karaman2011sampling}, \fmt \citep{janson2015fast}) approach the complexity of the motion planning problem by only implicitly representing the state space with a set of probing samples in $\xfree$ and making local, free connections to neighbor samples (sampled states within an $r_n>0$ cost radius, where $n$ is the number of samples).
These samples are drawn from a sample source (random or deterministic) and then distributed over the state space \citep{hsu2006probabilistic}.
As more samples are added, the implicit representation is able to model the true state space arbitrarily well, allowing theoretical guarantees of both completeness (a solution will be found, if one exists) and asymptotic optimality (the cost of the found solution converges to the optimum) \citep{karaman2011sampling,janson2018deterministic}.
This work focuses on computing sampling distributions to allocate samples to regions more likely to contain an optimal motion plan.

\section{Learning-Based Sample Distributions}\label{sec:method}

The goal of this work is to develop a methodology capable of identifying regions of the state space containing optimal trajectories with high probability, and generating samples from these regions to improve the performance of sampling-based motion planning (SBMP) algorithms.
These regions may be arbitrary and potentially complex, defined by internal or external factors.
Specifically, we refer to intrinsic properties of the robotic system, independent of the individual planning problem (e.g., the system dynamics) as internal factors, and we refer to properties specific to the planning problem itself (e.g., the obstacles, the environment, the initial state, and the goal region) as external factors.
At the core of our methodology is a Conditional Variational Autoencoder (CVAE), as it is expressive enough to represent very complex, high-dimensional distributions and general enough to admit arbitrary problem inputs. 
The CVAE is an extension of the standard variational autoencoder, which is a class of generative models that has seen widespread application in recent years \citep{kingma2013auto}. 
This extension allows conditional data generation by sampling from the latent space of the model \citep{sohn2015learning}; in the motion planning context, the conditioning variables represent external factors.
Lastly, these samples are used, along with uniform samples that ensure state space coverage, as the sampling distribution for SBMP algorithms.
The method thus leverages previous robotic experience (motion plans and demonstrations) to inform planning algorithms.
This combination of learning and SBMP allows both the generality of learning and the exhaustive exploration ability and theoretical guarantees of SBMP.

\begin{figure}[htb]
\centering
    \begin{subfigure}{0.5\textwidth}
        \includegraphics[width=\textwidth]{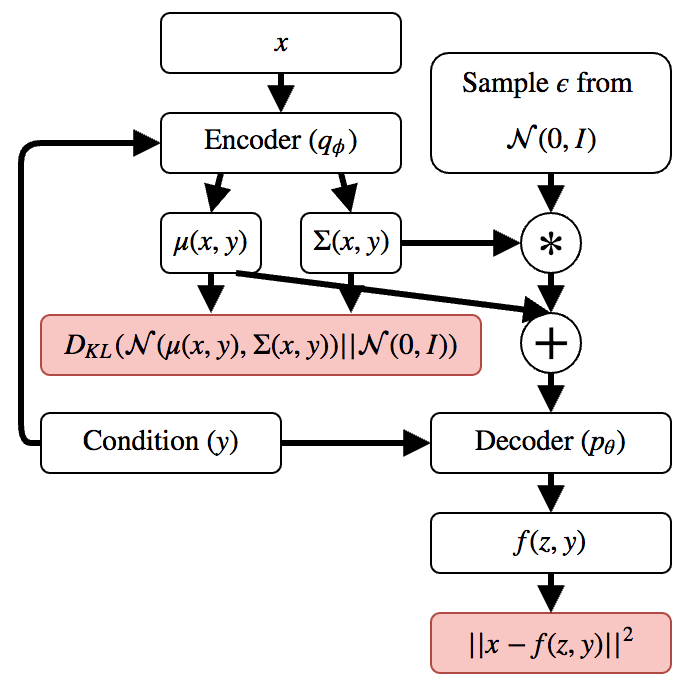}
        \caption{Training}
        \label{fig:cvaeTrain}
    \end{subfigure}
    \begin{subfigure}{0.4\textwidth}
        \includegraphics[width=\textwidth]{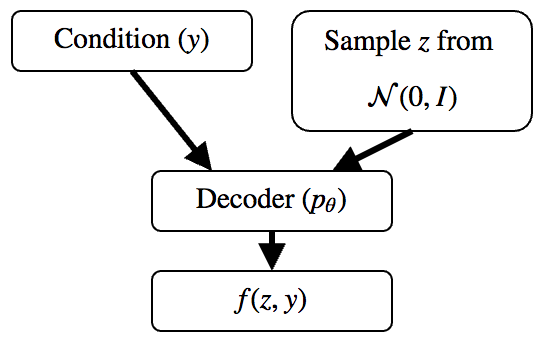}
        \caption{Sampling}
        \label{fig:cvaeSample}
    \end{subfigure} 
\caption{Conditional Variational Autoencoder (CVAE) setup \citep{doersch2016tutorial}. In the context of this work, $x$ represents training states, $y$ the conditioning variable (possibly initial state, goal region, and workspace information), and $z$ the latent state. The decoder (b) is used online to project conditioned latent samples into our distribution.
}
\label{fig:cvae}
\end{figure}

We will briefly discuss the theory behind the variational autoencoder (VAE), and point out connections to concrete features of our methodology. 
We aim to construct a distribution for the set of sampled points lying on a nearly optimal trajectory, conditioned on a given planning problem. 
We will refer to a sampled point as $x$.
We will denote a finite dimensional encoding of the planning problem and other external features as $y$. 
For example, a map of obstacles in a workspace can be encoded as an occupancy grid, for which $y$ is an array of binary elements. 
We will denote the conditional density of sample points, conditioned on $y$, as $p(x|y)$.
This distribution may be formulated as a latent variable model, where we write the joint density of sampled points and the latent variable as $p(x|z,y) \, p(z|y)$, where $z$ is a latent variable. 
We may then write parameterized forms of these densities as $p_\theta (x|z,y)$ and $p_\theta(z|y)$ respectively, where $\theta$ is a vector of parameters. 
Given this formulation, the maximum likelihood approach aims to maximize the likelihood
\begin{equation}
\label{eq:intrac}
    p_\theta(x) = \int p_\theta(x|z,y) \, p_\theta(z|y) dz
\end{equation}
with respect to the empirical distribution. In this work, as is standard in the VAE literature \citep{doersch2016tutorial}, we will let $p_\theta(x|z,y) = \mathcal{N}(x|f(z, y; \theta), \, \sigma^2 * I)$, where $\sigma^2$ is a hyperparameter that is set to be a small value, and $f$ is a deterministic function which will be encoded as a neural network (typically referred to as the decoder). 
Because any distribution over the latent variable may be mapped to an arbitrary distribution by the nonlinear function $f$, we will let $p_\theta(z|y) = \mathcal{N}(0,I)$.
However, computing the integral in (\ref{eq:intrac}) is intractable. 
To address this problem, the approach taken in variational inference is to approximate the posterior $p(z|x,y)$ with a function $q_\phi(z|x,y)$, where $\phi$ is a vector of parameters. This is referred to as the encoder. 
A divergence penalty is then enforced between $p(z|x,y)$ and $q_\phi(z|x,y)$.
With some manipulation, the log likelihood $ \log p_\theta(x|y)$ may then be written as
\begin{equation}
    \log p_{\theta}(x|y) - D_{KL}(q_{\phi}(z|x, y)\, \|\, p_{\theta}(z|x, y)) =  
    \mathbb{E}_{q_{\phi}(z|x, y)}[\log p_{\theta}(x|z,y)] - D_{KL}(q_{\phi}(z|x, y)\, \|\, p_{\theta}(z|y)),
\end{equation}
where $D_{KL}$ denotes the KL divergence. We refer the interested reader to \citep{kingma2013auto,doersch2016tutorial} for further details. The right hand side of this equation is referred to as the Evidence Lower Bound (or ELBO), as it is a lower bound on the log likelihood resulting from the non-negativity of the KL divergence. 
Because the KL divergence term on the right hand side is small (due in part to using high capacity models in the form of neural networks), we can optimize the right hand side as a tractable surrogate for the log likelihood. 
This is then optimized with respect to the parameters $\theta$ and $\phi$ via backpropagation. 
Writing $q_\phi(z|x,y) = \mathcal{N}(\mu(x,y),\,\Sigma(x,y))$, and noting $p_\theta(z|y)$ is modelled as an isotropic, unit-variance Gaussian, maximizing the log likelihood lower bound above is equivalent to maximizing
\begin{equation}
\label{eq:recon}
\| x - f(z,y)\|^2 - D_{KL}(\mathcal{N}(\mu(x,y),\Sigma(x,y)) \,\| \,\mathcal{N}(0,I))
\end{equation}
with respect to $\theta$ and $\phi$. To make this tractable via backpropagation, the reparameterization trick is used \citep{kingma2013auto}. Roughly, this is equivalent to modeling $q_\phi$ as a deterministic function with a stochastic input, such that $z = \mu(x,y) + A\,\epsilon$, where $\epsilon \sim \mathcal{N}(0,I)$ and $A A^T = \Sigma(x,y)$. The outline of the training process is provided in Fig.~\ref{fig:cvaeTrain}. The optimization of Equation~\ref{eq:recon} is done via standard stochastic gradient methods.

The standard construction of the conditional VAE (CVAE) consists of neural networks for the encoder $q_\phi(z|x,y)$ and the decoder $p_\theta(x|z,y)$. 
Once trained, the decoder allows us to approximately generate samples from $p(x|y)$ by simply sampling from the normal distribution of the latent variable $p(z|y) = \mathcal{N}(0,I)$ (see Fig.~\ref{fig:cvaeSample}). 
While one iteration of this offline phase is often sufficient, with problems that are expensive to solve and thus expensive to acquire data for, the entire methodology may be performed iteratively. 
Thus, a partially trained CVAE may generate samples that result in better planning performance, allowing more, high-quality data and subsequently allowing the CVAE to be further trained.
In practice, it is common to add a weighting term ($\beta$) to the KL divergence term in the ELBO \citep{higgins2017beta}. 
This term controls the relative weighting of the autoencoding loss (the reconstruction error) and the strength of the prior over $z$ \citep{alemi2018fixing}. 
The value of $\beta$ was chosen on a per-problem basis.

\paragraph{Approach.} We now examine the methodology in detail, following along with the outline below.
It begins with an offline phase which trains the CVAE, to be later sampled from.
Line \ref{line:inputData} initializes this phase with the required demonstration data. 
This data (states and any additional planning problem information) may be from successful motion plans, previous trajectories in the state space, human demonstration, or other sources that provide insight into how the system operates. 
In this work we use each of these data sources (Section \ref{sec:numexp}), though, when available, optimal solutions to previous motion planning problems are preferred since these will intuitively provide the most insight into the optimal motion planning problem.
In order to generate the required breadth of data (in this work, on the order of one hundred thousand data points), we leverage GPU-accelerated, approximate motion planning algorithms to generate plans quickly \citep{ichter2017group}.
The data is then processed into the state of the robot and the conditioning variables.
In particular, these conditioning variables (Line \ref{line:cond1}) contain information about the problem, such as workspace information (e.g., obstacles) or the initial state and goal region. 
The CVAE is then trained in Line \ref{line:train}, with the goal of learning the internal representation of the system conditioned on external properties of the problem (which may inform where in the state space the system will operate, adaptively to a problem).

\begin{algorithm}[htb]
\caption*{\textbf{Learning Sample Distribution Methodology Outline}}
\label{alg:outline}
\algsetup{linenodelimiter=}
\begin{algorithmic}[1]
\REQUIRE 
\STATE \textbf{Input:} Data (successful motion plans, robot in action, human demonstration, etc.) \label{line:inputData}
\STATE Construct conditioning variables $y$ \label{line:cond1}
\STATE Train CVAE, as in Fig.~\ref{fig:cvaeTrain} \label{line:train}
\ENSURE 
\STATE \textbf{Input:} New motion planning problem $(\xfree, \xinit, \xgoal)$, learned sample fraction $\lambda$ \label{line:inputsMPP}
\STATE Construct conditioning variable $y$ \label{line:cond2}
\STATE Generate $\lambda N$ free samples from the CVAE latent space conditioned on $y$, as in Fig. \ref{fig:cvaeSample}\label{line:sampleLatent}
\STATE Generate $(1-\lambda)N$ free samples from an auxiliary (uniform) sampler \label{line:sampleAux}
\STATE Run sampling-based planner (e.g., \prm, \fmt, \rrt)\label{line:run}
\end{algorithmic}
\end{algorithm}

The online phase of the methodology begins with a new planning problem, Line \ref{line:inputsMPP}, defined by the tuple $(\xfree, \xinit, \xgoal)$, which is formed into a conditioning variable $y$ in Line \ref{line:cond2}. For example, $y$ may be the initial state, the goal region, or workspace obstacles encoded in an occupancy grid.
With this in hand, we now generate samples by sampling the latent space as  $\mathcal{N}(0,I)$, conditioning on $y$, and mapping these samples to the state space through the decoder network (Line \ref{line:sampleLatent}).
In order to maintain the ability of SBMP algorithms to represent the true state space with arbitrarily high fidelity, and thus maintain the theoretical guarantees of SBMP algorithms (see Remark \ref{rem:pcao}), we also sample from an auxiliary sampler, in our case a uniform sampler.
We denote the fraction of learned samples as $\lambda$, i.e., we generate $\lambda N$ samples from the learned sampler and $(1-\lambda)N$ from the auxiliary sampler.
We have found through experimentation (Section \ref{sec:lambda}) that $\lambda = 0.5$ represents a satisfactory balance between leveraging the learned sample regions and ensuring full coverage of the state space.
In particular, the learned sampler is often able to find solutions quickly, with very few samples. 
However, if the learned sampler does not fully identify the region containing the optimal solution, the uniform sampler must effectively fill in the gaps, i.e., the learned sampler will continue to miss these regions even with more samples.
Finally, in Line \ref{line:run}, we use these samples to seed a SBMP algorithm, such as \prm, \fmt, or \rrt, and solve the planning problem. This methodology is applied to a variety of problems with varying state space dimensionality, constraints, and training data-generation approaches in the following section. 

\begin{remark}[Probabilistic Completeness and Asymptotic Optimality] \label{rem:pcao}
Note that the theoretical guarantees of probabilistic completeness and asymptotic optimality from \citep{janson2015fast}, \citep{janson2018deterministic}, and \citep{karaman2011sampling} hold for this method by adjusting any references to $n$ (the number of samples) to $(1-\lambda)N$ (the number of uniform samples in our methodology).
This result is detailed in Appendix D of \citep{janson2015fast} and Section 5.3 of \citep{janson2018deterministic}, which show that adding samples can only improve the solution or lower the dispersion (which the theoretical results are based on), respectively.
\end{remark}

\section{Numerical Experiments: Performance and Scalability}\label{sec:numexp}

To demonstrate the performance and generality of learning sample distributions, this section shows several numerical experiments with a variety of robotic systems.
The results in Section \ref{sec:geometric} were implemented in MATLAB with the Fast Marching Tree (\fmt) and Batch Informed Trees (\bit) algorithms \citep{janson2015fast,gammell2015batch}, while the remainder of the results were implemented in CUDA C with a GPU version of the Probabilistic Roadmap (\prm) algorithm (for convergence plots) and the Group Marching Tree (\gmt) algorithm (to generate training data) \citep{karaman2011sampling,ichter2017group}.
The CVAE was implemented in TensorFlow.
The simulations were then run on a Unix system with a 3.4 GHz CPU and an NVIDIA GeForce GTX 1080 Ti GPU.
Example code and the network architecture may be found at \url{https://github.com/StanfordASL/LearnedSamplingDistributions}.
We begin with a simple geometric planning problem in which we show our method performs as well as or better than state of the art approaches. We also note that these state of the art approaches are less general than the method we present in this paper, and tuned well towards these geometric problems.
We then demonstrate the benefits of learning distributions for a high-dimensional spacecraft system, a dynamical system conditioned on workspace obstacle information, and a kinematic chain.
These results are examined conceptually as well as quantitatively in terms of convergence, finding an order of magnitude improvement in success rate and cost.

This section aims to show that the method presented in this paper achieves good performance on a wide variety of systems, from simple to complex. Moreover, this section examines a variety of conditioning variables, showing the approach is useful with no conditioning information (and the approach learns to sample based on characteristics of the system dynamics) or with complicated conditioning variables such as workspace representations. 
Note sample generation time is included in runtime, but generally accounts for only a fraction of the total runtime--generating thousands of samples takes only few milliseconds. 
The offline portion of training is not included in the runtime and was on the order of several minutes (see video at \url{https://goo.gl/E3JPWn} for training times with the problem in Section \ref{sec:workspace}).

\subsection{Geometric Planning Comparisons}\label{sec:geometric}

\begin{figure}[htb]
\centering
        \begin{subfigure}{0.31\textwidth}
        \includegraphics[width=\textwidth]{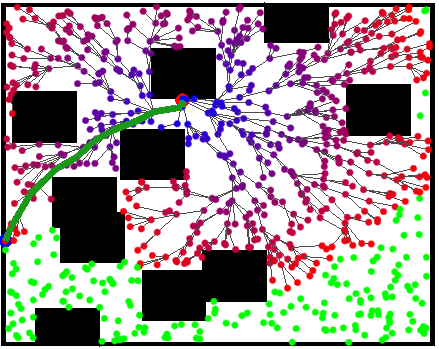}
        \caption{Uniform FMT$^\ast$}
        \label{fig:treeFMT}
    \end{subfigure} 
        \begin{subfigure}{0.31\textwidth}
        \includegraphics[width=\textwidth]{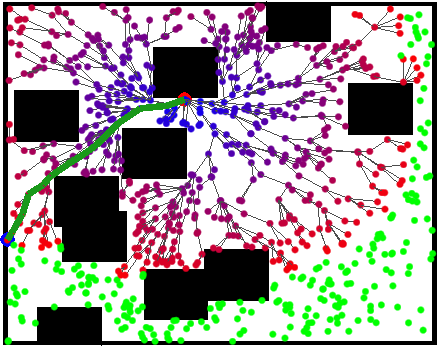}
        \caption{Hybrid FMT$^\ast$}
        \label{fig:treeHyFMT}
    \end{subfigure} 
    \begin{subfigure}{0.31\textwidth}
        \includegraphics[width=\textwidth]{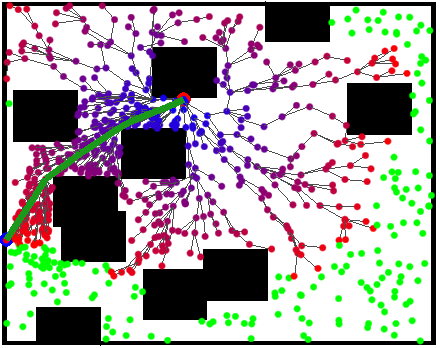}
        \caption{Learned FMT$^\ast$}
        \label{fig:treeLearnFMT}
    \end{subfigure} 
    \begin{subfigure}{0.45\textwidth}
        \includegraphics[width=\textwidth]{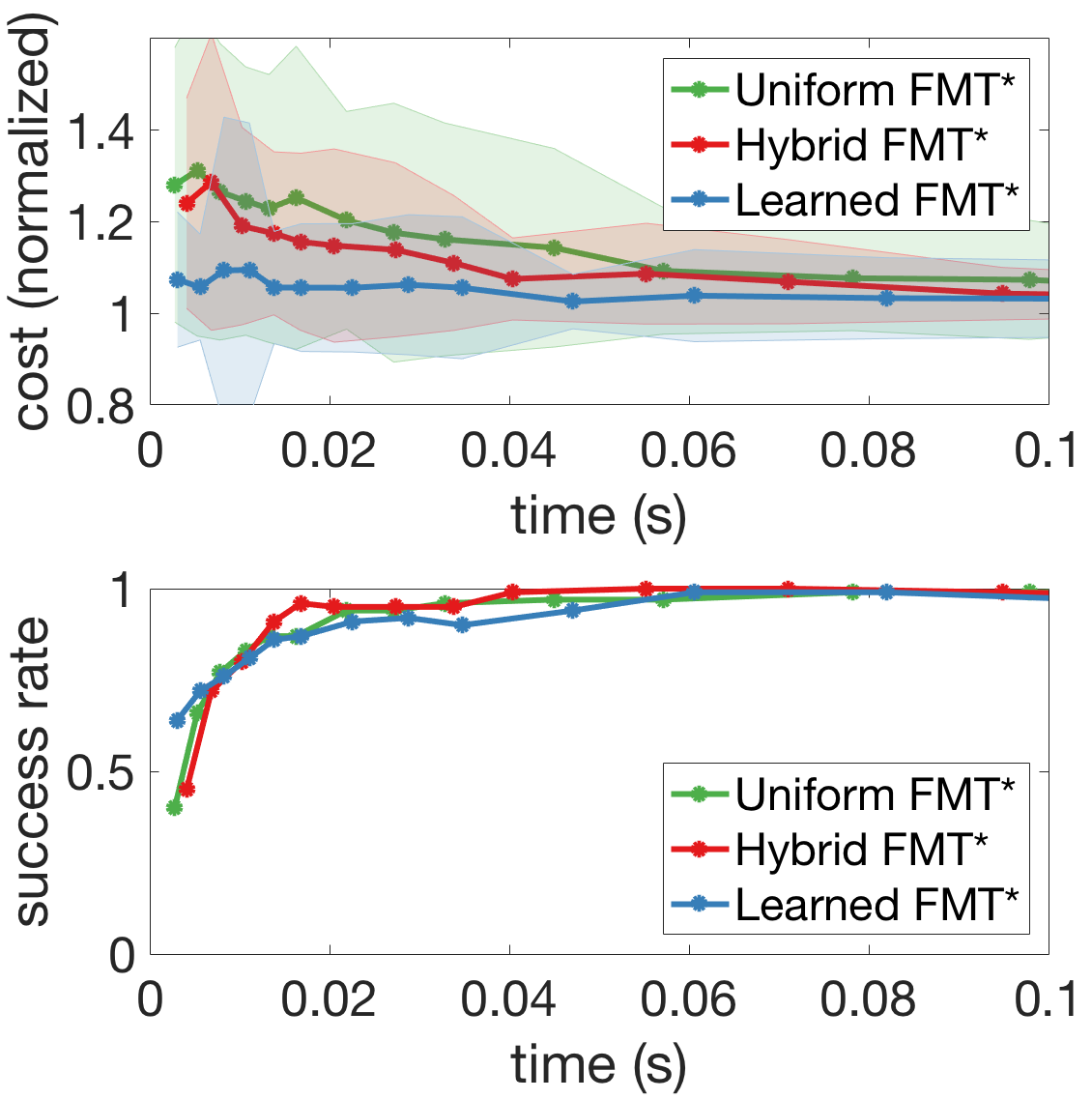}
        \caption{Convergence for \fmt}
        \label{fig:geometricConvergence}
    \end{subfigure}
    \begin{subfigure}{0.45\textwidth}
        \includegraphics[width=\textwidth]{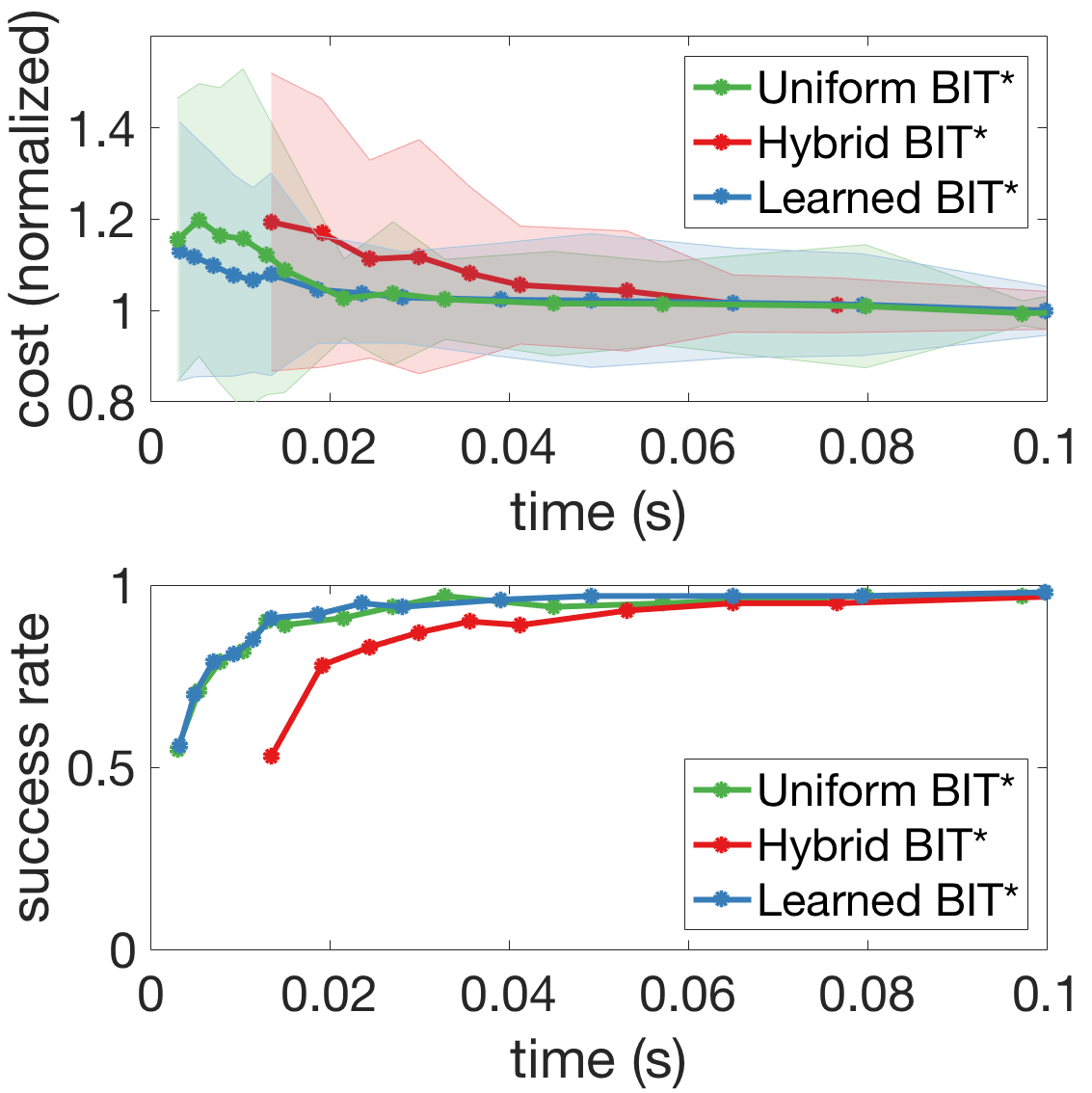}
        \caption{Convergence for \bit}
        \label{fig:geometricConvergenceBIT}
    \end{subfigure}
\caption{(\ref{fig:treeFMT}-\ref{fig:treeLearnFMT}) Solutions to the geometric planning problem with different sample distributions (colored by cost to come, or green if unexplored) and (\ref{fig:geometricConvergence}-\ref{fig:geometricConvergenceBIT}) convergence results for sample distributions with \fmt and \bit (results averaged over 100 runs, standard deviation shaded, and $\lambda = 0.5$). \textit{Hybrid} refers to the sampling strategy of \citep{hsu2005hybrid}, and \textit{Learned} refers to the method we present in this paper. 
}
\label{fig:geometricResults}
\end{figure}

While this methodology is very general and can be applied to complex systems,  we first show the methodology performs well for a simple, geometric problem.
All problems are created with randomly generated initial states, goal regions, and 10 cube obstacles, as shown in Fig.~\ref{fig:geometricResults}.
The learned sample distributions were conditioned on all the problem information (initial state, goal region, and obstacles), and trained over successful motion plans.

For this problem, we make comparisons to a non-uniform sampling strategy and combine our methodology with an exploration-guided non-uniform sampling algorithm.
Specifically, we consider the hybrid sampling strategy proposed in \citep{hsu2005hybrid}, and Batch Informed Trees (\bit) \citep{gammell2015batch}.
The hybrid sampling approach uses uniform samples, Gaussian samples, and bridge samples to create a distribution favoring narrow passageways and regions nearby obstacles.
\bit uses successive batches of samples to iteratively refine a tree, leveraging solutions from previous batches to selectively sample only states that can improve the solution.

The results of these comparisons are shown in Fig.~\ref{fig:geometricConvergence}.
The first comparison shows each strategy with \fmt \citep{janson2015fast}.
We find that each strategy performs nearly equivalently in terms of finding a solution, but the learned strategy finds significantly better solutions in the same amount of time. 
In fact, the learned sampling strategy finds within 5\% of the best solution almost immediately, instead of converging to it as the number of samples increase.
The results show less of a performance gap with \bit, though the learned strategy continues to perform at least as well as the others.
The delayed start of the hybrid convergence is only due to the time required to generate samples, which for \bit was implemented here by rejection.

\subsection{Spacecraft Debris Recovery}\label{sec:spacecraft}

\begin{figure}[htb]
\centering
	\includegraphics[width=0.6\textwidth]{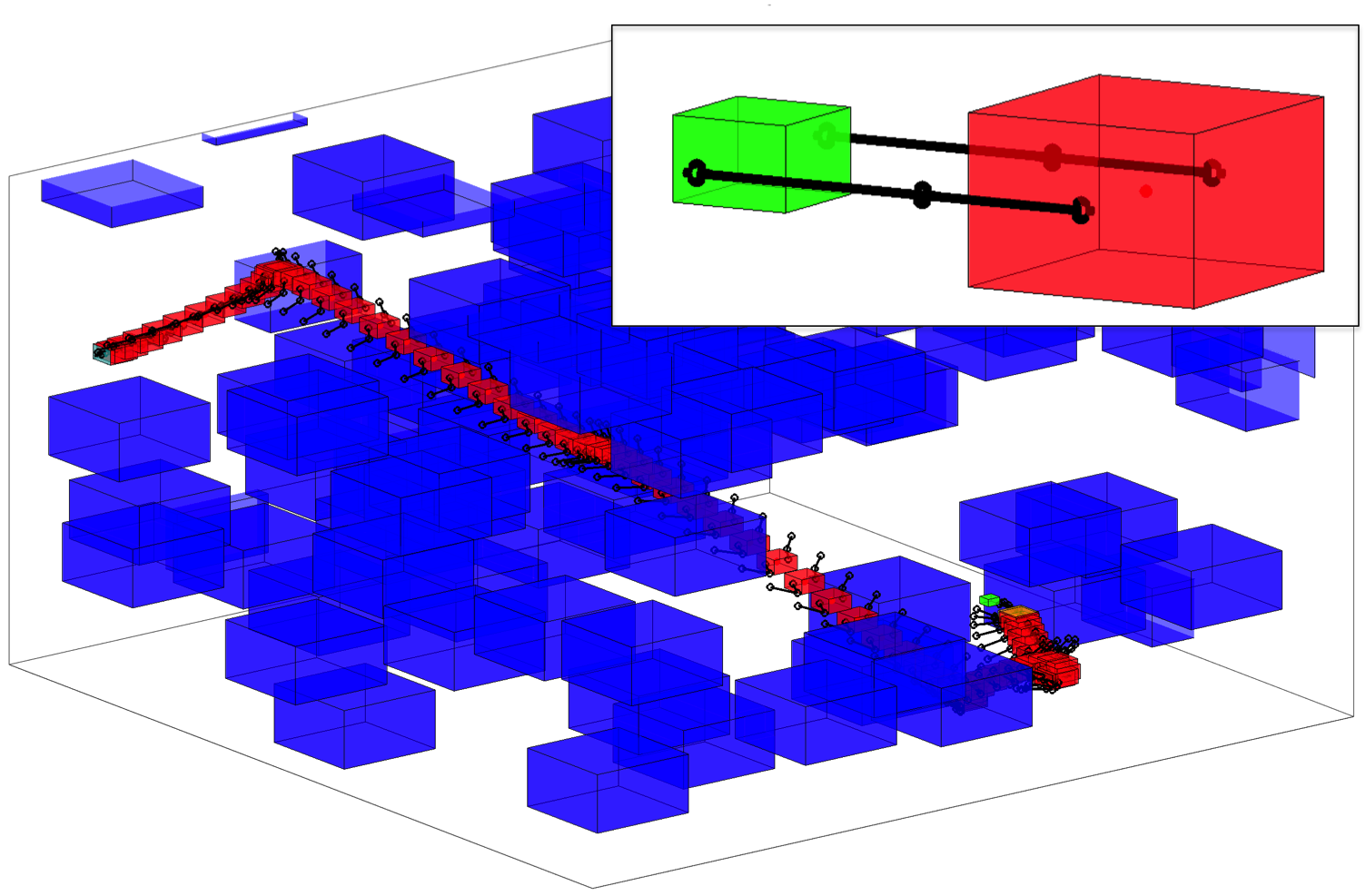}
	\caption{An example spacecraft debris recovery problem, whereby the spacecraft must maneuver from an initial state to recover debris (shown in the figure inset in green) between its end effectors, while avoiding obstacles (blue). The spacecraft (red) is modeled as a double integrator with a pair of 3 DoF kinematic arms (shown in the figure inset in black).}
\label{fig:spacecraftProblem}
\end{figure}

The second numerical experiment considered is a simplified spacecraft debris recovery problem, whereby a cube shaped spacecraft with 3D double integrator dynamics ($\ddot{x} = u$, no rotations) and a pair of 3 DoF kinematic arms (assumed to be much less massive than the spacecraft body), for a total of 12 dimensions, must maneuver from an initial state, through a cluttered asteroid field, to recover debris between its end effectors.
The cost is set as a mixed time/quadratic control effort penalty with an additional term for joint angle movement in the kinematic arms.
The initial states and goal regions were randomly generated, as were the asteroids (i.e., obstacles).
Fig.~\ref{fig:spacecraftProblem} shows an example problem and the spacecraft setup.
The CVAE was conditioned on the initial state, goal region, and debris location, and trained with successful motion plans.

\begin{figure}[htb]
\centering
    \begin{subfigure}{0.48\textwidth}
        \includegraphics[width=\textwidth]{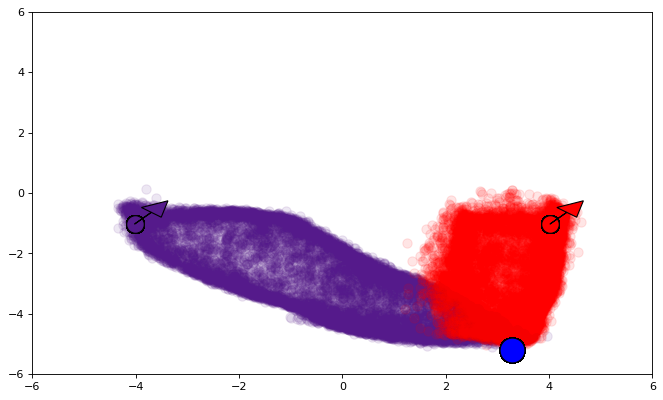}
        \caption{$x$ vs $y$}
        \label{fig:spacecraftDistXY}
    \end{subfigure} 
    \begin{subfigure}{0.48\textwidth}
        \includegraphics[width=\textwidth]{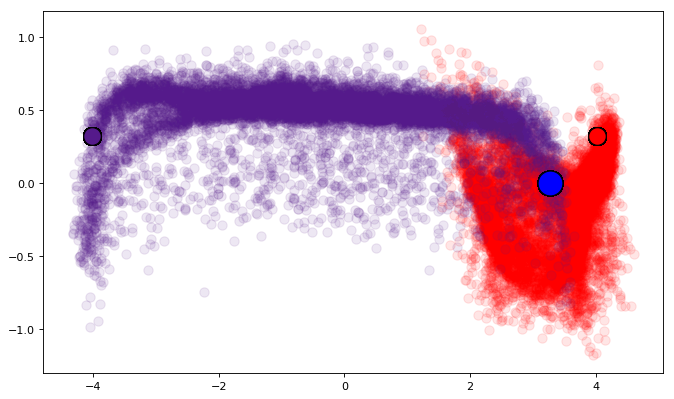}
        \caption{$x$ vs $\dot{x}$}
        \label{fig:spacecraftDistXXdot}
    \end{subfigure} 
    \begin{subfigure}{0.48\textwidth}
        \includegraphics[width=\textwidth]{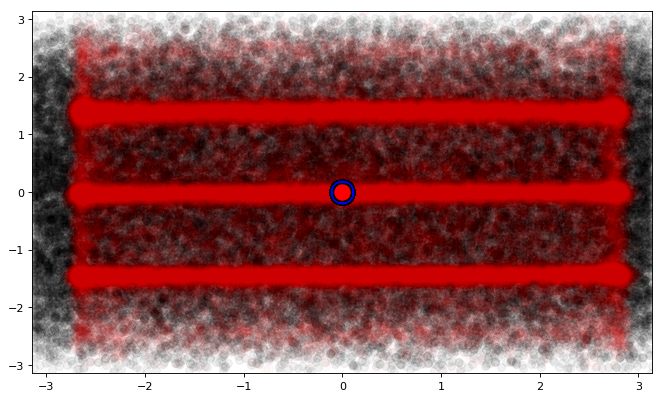}
        \caption{$\alpha$ vs $\beta_1$}
        \label{fig:spacecraftDist12}
    \end{subfigure} 
    \begin{subfigure}{0.48\textwidth}
        \includegraphics[width=\textwidth]{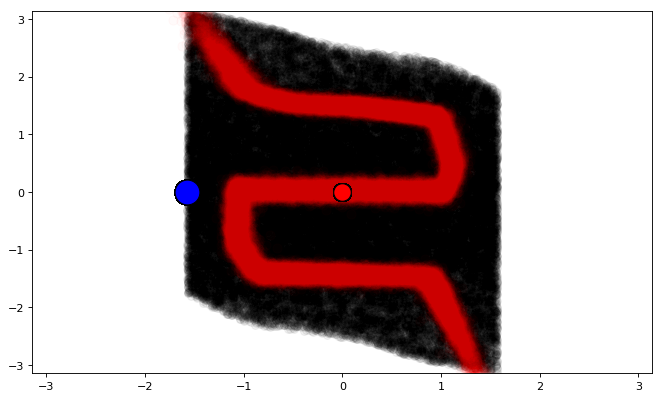}
        \caption{$\beta_1$ vs $\beta_2$}
        \label{fig:spacecraftDist23}
    \end{subfigure} 
\caption{Spacecraft learned distributions for various dimensions. The learned distributions are shown in red and purple (corresponding to different initial states, plotted as red and purple circles), the goal region is shown in blue, and the initial training distributions are shown in black (when displayed). (\ref{fig:spacecraftDistXY}) shows the distribution favoring an ellipse connecting the initial state and goal region. (\ref{fig:spacecraftDistXXdot}) shows a phase portrait of the $x$ dimension, where the samples favor velocities towards the goal region. (\ref{fig:spacecraftDist12}-\ref{fig:spacecraftDist23}) show distributions for the kinematic arm angles, where it has converged to a few fixed values to sample, thus reducing movement cost ($\alpha$ and $\beta_1$ denote rotation angles at the arm-spacecraft hub and $\beta_2$ denotes the joint angle in the arm). 
}
\label{fig:spacecraftDistributions}
\end{figure}

\begin{figure}[htb]
\centering
       \includegraphics[width=0.7\textwidth]{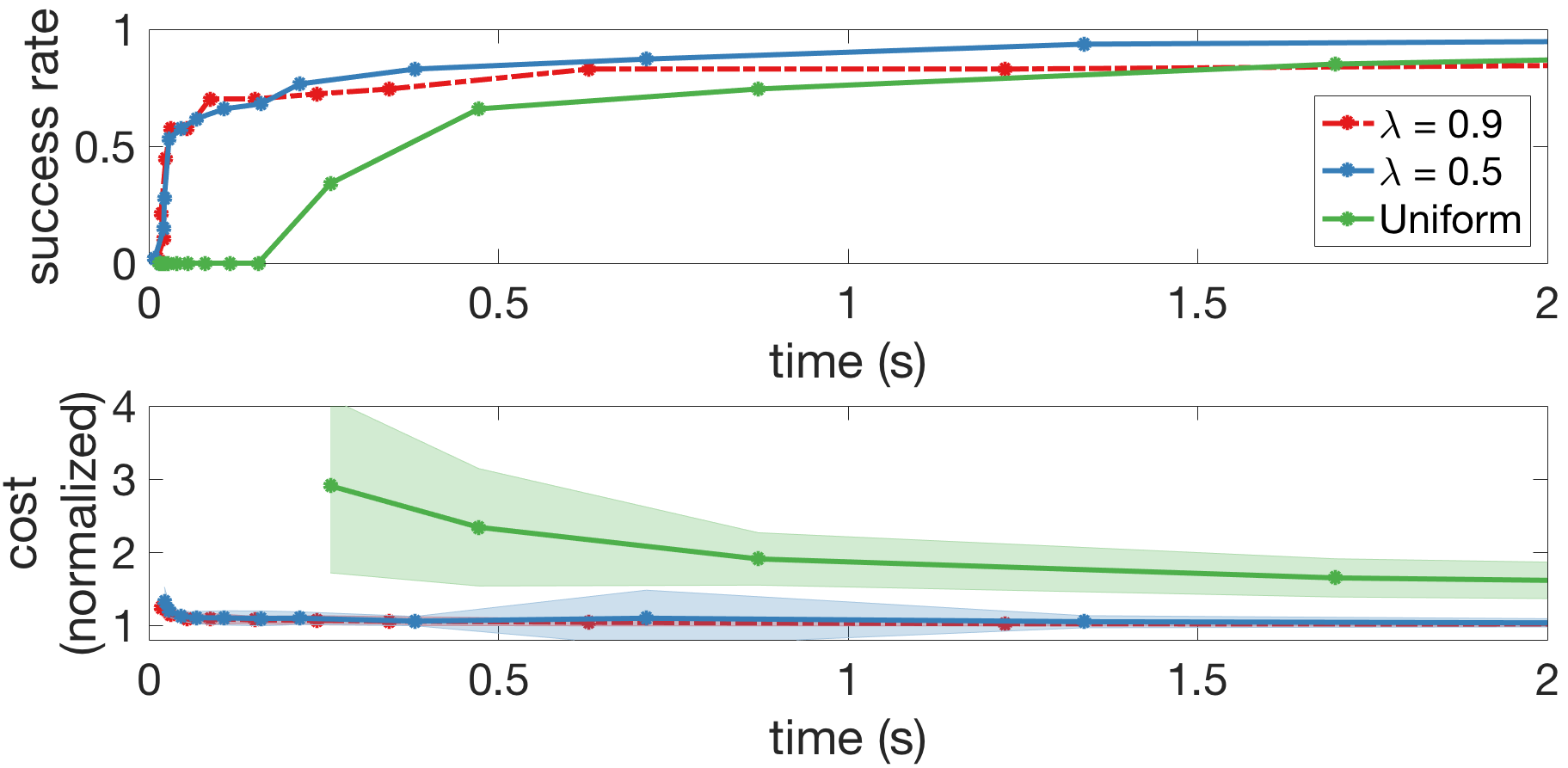}
 \caption{Convergence results for the spacecraft planning problem. Planning with the learned distributions (50\% and 90\% learned) significantly outperforms planning with a uniform distribution (results averaged over 100 runs and the standard deviation shaded).
}
\label{fig:spacecraftConvergence}
\end{figure}

The resulting learned distributions are shown in Fig. \ref{fig:spacecraftDistributions}.
Fig. \ref{fig:spacecraftDistXY} shows the learned distribution of the $x$ and $y$ positions  for two problems. 
The distribution resembles an ellipsoid connecting the initial state and goal region, with some spread in the minor axis to account for potential obstacles in the trajectory and a slight skew in the direction of the initial velocity--we note the similarity to the sample distributions found after exploration in \bit \citep{gammell2015batch}.
Fig. \ref{fig:spacecraftDistXXdot} shows the learned distribution of $x$ and $\dot{x}$, i.e., a phase portrait of the $x$ dimension. 
The purple distribution favors velocities such that any sample flows from the initial state to the goal region, first accelerating near the maximum sampled velocity ($\dot{x} = 1$), and then maintaining the velocity until nearby the goal.
The red distribution, whose initial position is much closer to the goal region has a much larger spread, favoring samples with velocities towards the goal in all directions.
Finally, the learned distributions for a single arm are shown in Figs. \ref{fig:spacecraftDist12}-\ref{fig:spacecraftDist23}.
The angle distributions demonstrate the arm movement should be kept to a minimum, by holding one dimension fixed to a few values only.
In the problem setup, the arms have significantly less impact on obstacle avoidance, but can incur a large cost for movement, which is reflected in the distributions.

Fig. \ref{fig:spacecraftConvergence} shows the convergence of the methods in time. 
The learned sampling distribution outperforms the uniform by approximately an order of magnitude in finding solutions when they exist.
The cost convergence curves show that planning with learned samples converges almost immediately to within a few percentage of optimal, while even after 10,000 samples, planning with a uniform distribution is still more than 60\% from optimal. 
This immediate convergence is similar to what was observed in the geometric planning problem and the narrow passage problem (in the following section).
We also note the variance is smaller for the learned distributions.

\subsection{Workspace Learning}
\label{sec:workspace}

\begin{figure}[htb]
\centering
    \includegraphics[width=0.95\textwidth]{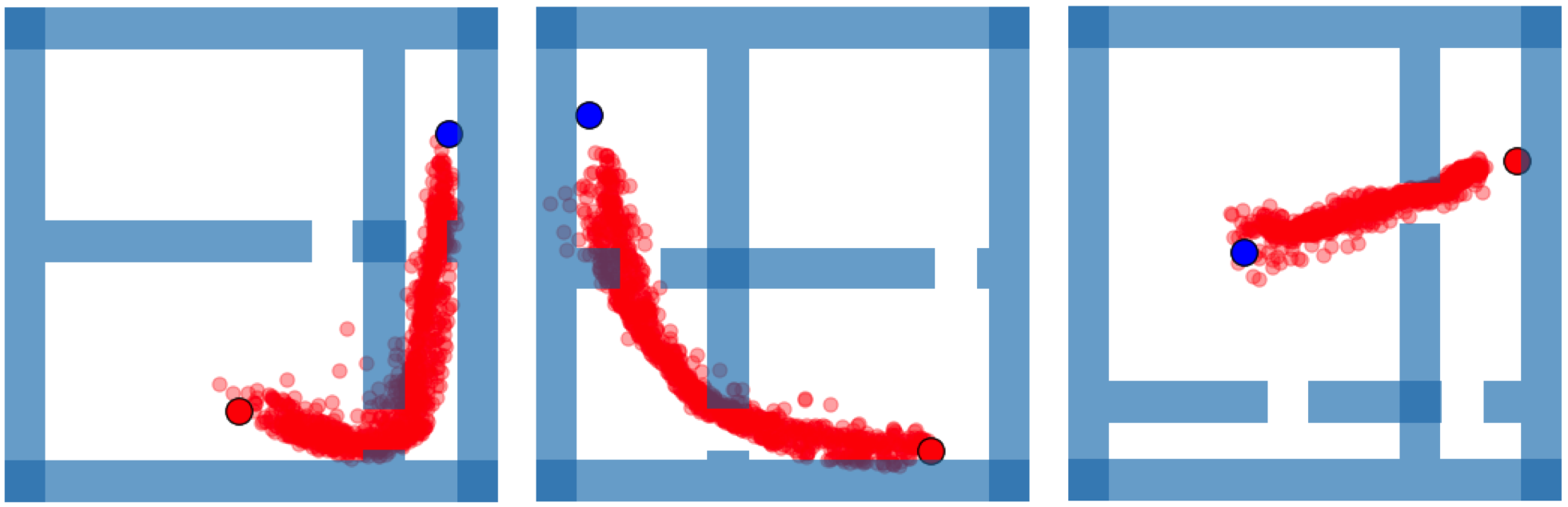} 
    \caption{Example learned distributions for the narrow passage problem, conditioned on the initial state (red), goal region (blue), and the obstacles (through an occupancy grid).
}
\label{fig:narrowPassageProblem}
\end{figure}

\begin{figure}[htb]
\centering
    \includegraphics[width=0.8\textwidth]{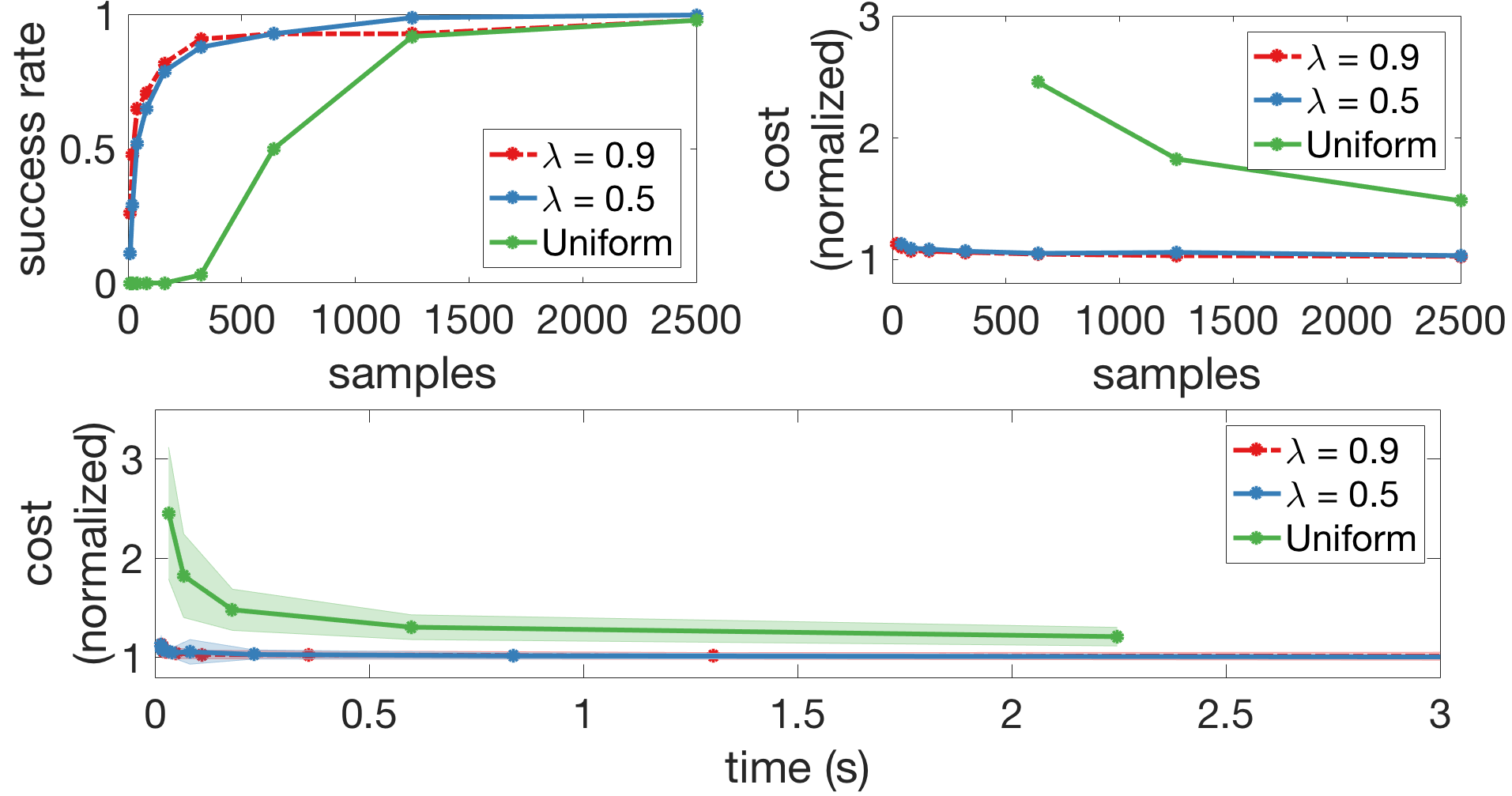}
    \caption{Convergence results for the narrow passage problem, demonstrating the learned sample distributions (50\% and 90\% learned) achieve approximately an order of magnitude better performance in terms of success rate, and are able to converge with few samples (results averaged over 100 runs and the standard deviation shaded).}
\label{fig:narrowPassageConvergence}
\end{figure}

The next problem, shown in Figs. \ref{fig:learnedTreeNarrow} and \ref{fig:narrowPassageProblem}, was loosely inspired by the narrow passage problems in \citep{zucker2008adaptive}, and demonstrates the ability of the methodology to learn distributions conditioned on workspace information.
The problem features a 3D double integrator (6 dimensional state space) operating in an environment with 3 narrow passages.
The initial state, goal region, and gap locations are all randomly generated and used to condition the CVAE for each problem (an occupancy grid was used to represent the obstacles in the conditioning variable).
Fig. \ref{fig:narrowPassageProblem} shows several problems and their learned distribution; clearly, the CVAE has been able to capture both initial state and goal region biasing, some sense of dynamics, and the obstacle set.
The velocity distributions too show the samples effectively favoring movement from the initial state to goal region.
The convergence results, shown in Fig. \ref{fig:narrowPassageConvergence}, demonstrate learned distribution solutions can be found with approximately an order of magnitude fewer samples and converge in cost almost immediately, while the uniform sampling results show the classic convergence curve we may expect. A video of the methodology applied to this problem can be found here, \url{https://goo.gl/E3JPWn}.

This method's ability to learn both dynamics and obstacles demonstrates that learning is capable of almost entirely solving some problems.
While this would be quite efficient, we also found the learned distributions susceptible to failure modes (e.g., cutting corners), which result in infeasible solution trajectories. 
In our methodology, this is easily handled through the uniform sampling and the guarantees of SBMP.
This methodology may thus be thought of as attempting to solve the problem through learning, and accounting for possible errors with a theoretically sound algorithm.

\subsection{Chain}

\begin{figure}[h!]
\centering
    \begin{subfigure}{0.38\textwidth}
        \includegraphics[width=\textwidth]{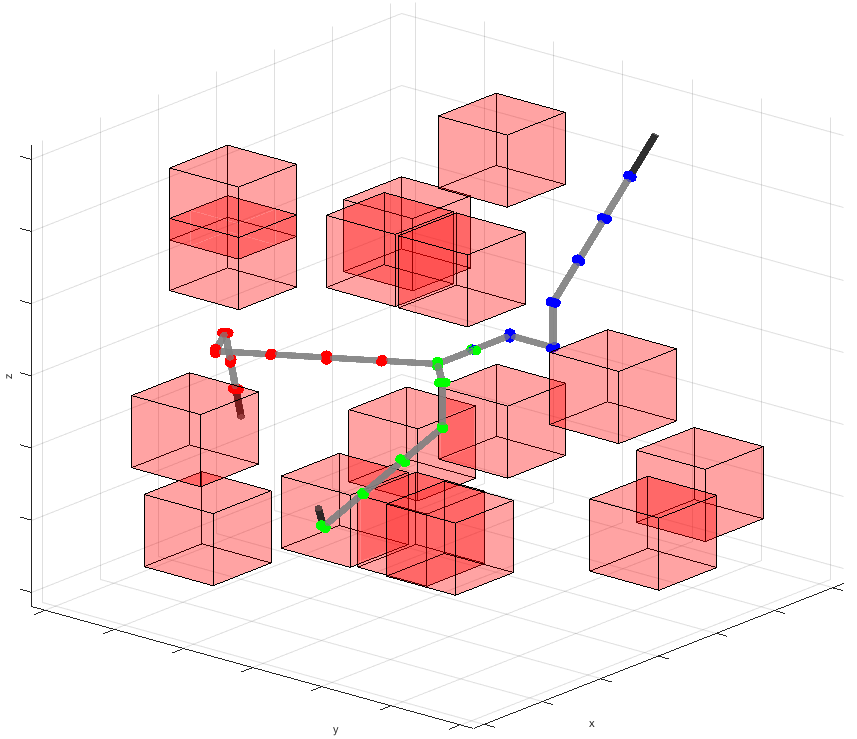}
        \caption{}
        \label{fig:chains}
    \end{subfigure} 
    \begin{subfigure}{0.38\textwidth}
        \includegraphics[width=\textwidth]{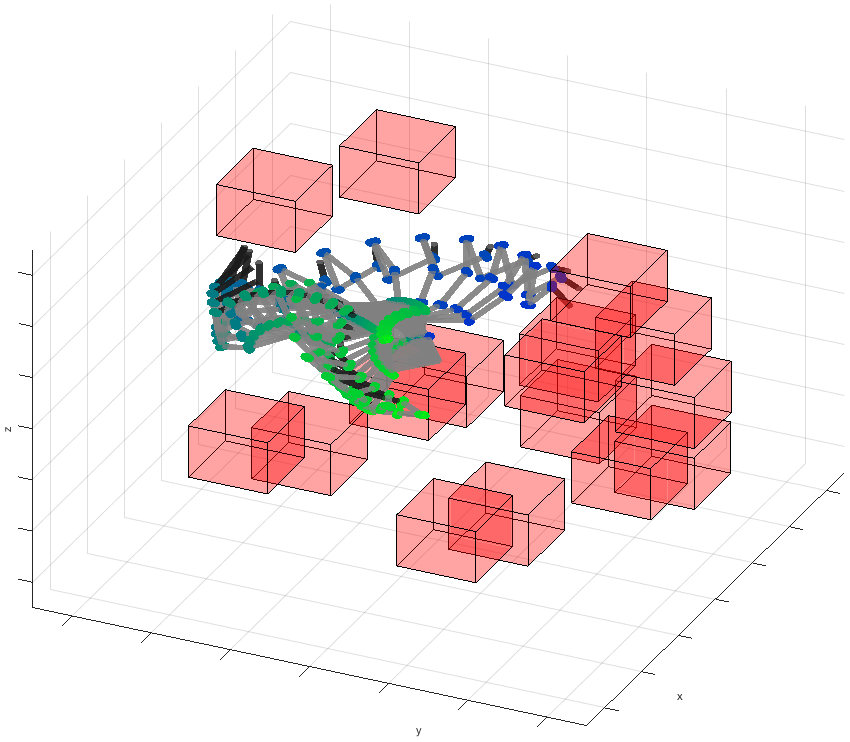}
        \caption{}
        \label{fig:chainSoln}
    \end{subfigure} 
    \begin{subfigure}{0.8\textwidth}
    	\centering
        \includegraphics[width=\textwidth]{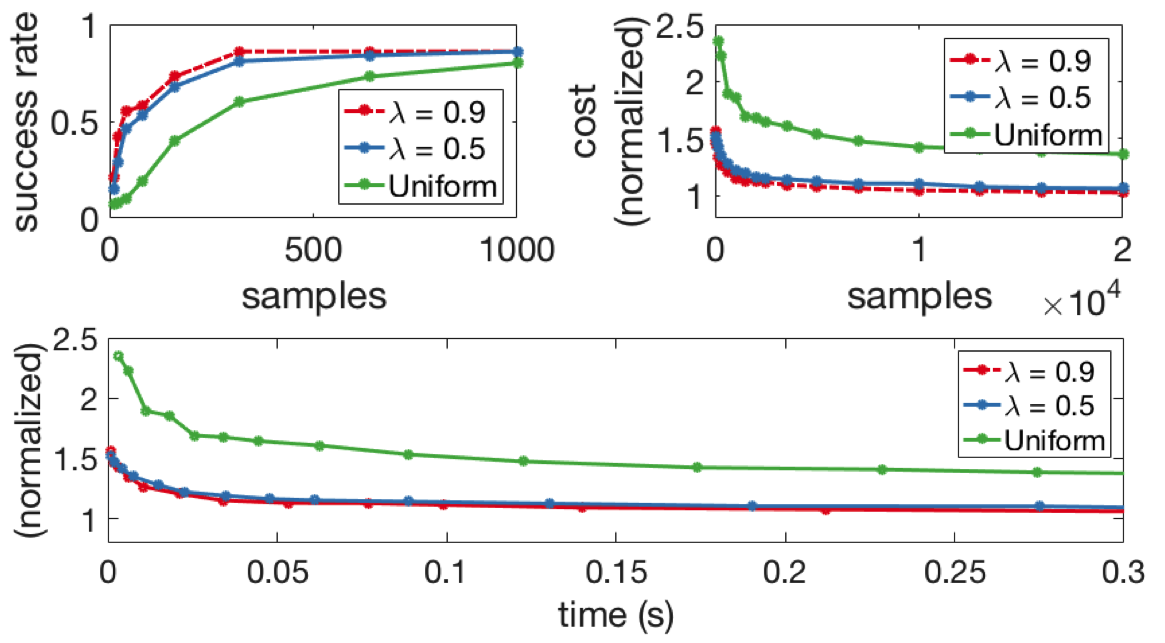}
        \caption{}
        \label{fig:chainConvergence}
    \end{subfigure} 
\caption{Results for the kinematic arm problem. (\ref{fig:chains}) shows three valid configurations of the system. (\ref{fig:chainSoln}) shows a successful trajectory. (\ref{fig:chainConvergence}) shows the performance in terms of the success rate and the cost, plotted versus the number of samples and the planning time. Within approximately 25ms, the performance of the learned sampling approach exceeds the performance of the uniform sampling approach after 300ms. 
}
\label{fig:chainResults}
\end{figure}

We next demonstrate our methodology on a kinematic arm planning problem. 
The arm, shown in three potential configurations in Fig. \ref{fig:chains}, has eight degrees of freedom.
Each degree of freedom is a rotational joint around an alternating axis.
The arm must navigate a cluttered environment from an initial state to a goal region as in Fig. \ref{fig:chainSoln}.
This scenario demonstrates a planning problem in which the optimal sample placement is unintuitive in the state space.
Still, the convergence results show similar performance increases.

\section{Numerical Experiments: Extensions, Data Sources, Generalization, and Hyperparameter Selection}
\label{sec:gen}

In this section, we investigate modifications to the learned sampling distribution methodology that can result in performance improvements. 
We first investigate the role of algorithmic parameters on the performance of the learned sampling distribution methodology. 
We investigate learning structured distributions in which samples are coupled together resulting in improved dispersion along the trajectory. 
We investigate out-of-distribution generalization.
Finally, we investigate potential training data sources when solution trajectories are not available.

\subsection{Fraction of Learned Samples ($\lambda$)}\label{sec:lambda}

\begin{figure}[htb]
\centering
    \begin{subfigure}{0.8\textwidth}
        \includegraphics[width=\textwidth]{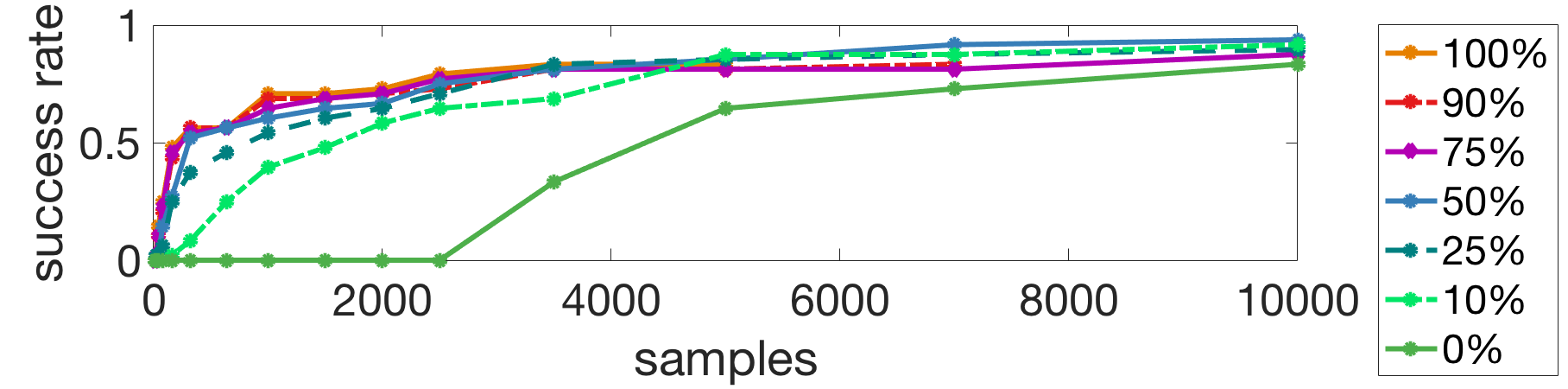}
        \label{fig:spacecraftConvergenceRatioNumSuccess}
    \end{subfigure}
    \begin{subfigure}{0.8\textwidth}
        \includegraphics[width=\textwidth]{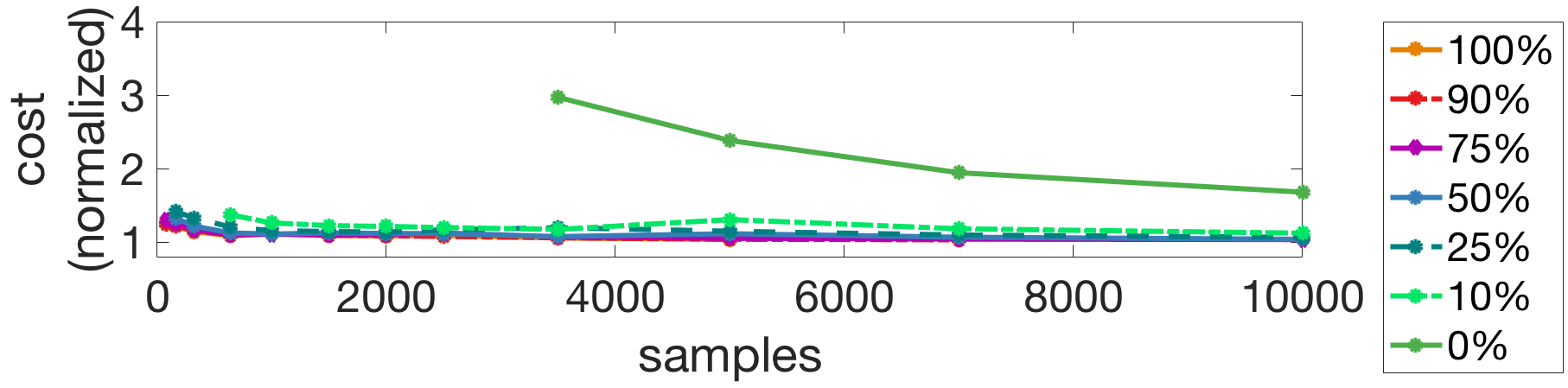}
        \label{fig:spacecraftConvergenceRatioNumCost}
    \end{subfigure}
    \begin{subfigure}{0.8\textwidth}
        \includegraphics[width=\textwidth]{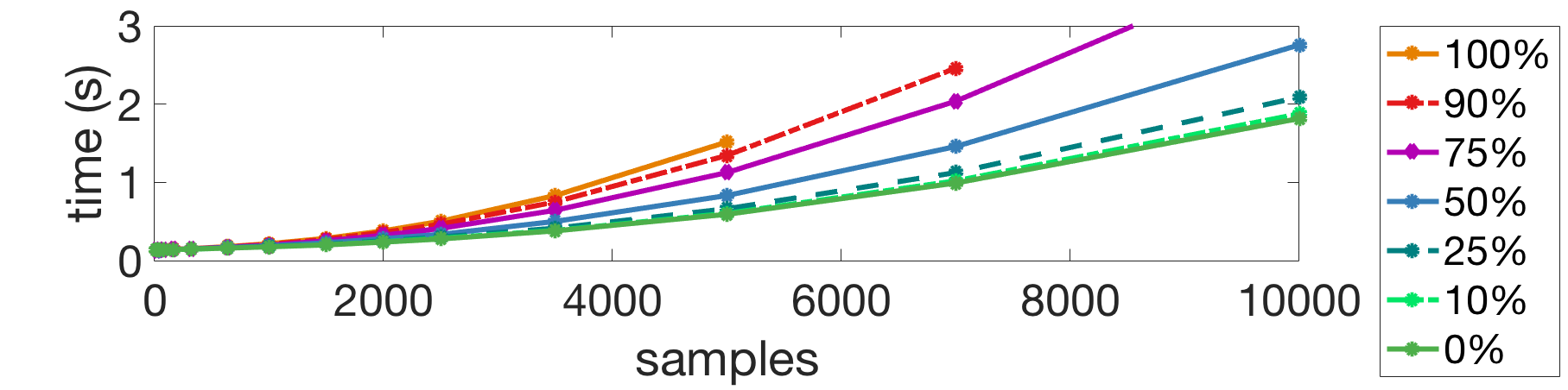}
        \label{fig:spacecraftConvergenceRatioNumTime}
    \end{subfigure}
\caption{Convergence comparison over varying percentages of learned samples to uniform samples (i.e., $\lambda$) in the spacecraft planning problem. From these results a 50-50 split was selected as an ideal in terms of runtime, success rates, and cost.
}
\label{fig:spacecraftConvergenceRatio}
\end{figure}

In this section we investigate the effect of the fraction of learned samples ($\lambda$) on the cost, time, and success rates. 
These comparisons are performed on the spacecraft environment (Section \ref{sec:spacecraft}).
Percentages between 0\% (all uniform) and 100\% (all learned) are shown in Fig. \ref{fig:spacecraftConvergenceRatio}.
In terms of convergence, all the percentages equal to or greater than 25\% performed equally well.
In terms of success rate, with small sample counts ($< 5000$), 50\% and above each performed similarly, however, as the number of samples increased, the high percentages (75\% and above) continued to fail on a few problems in which the learned sample distributions missed important regions of the state space.
Lastly, comparing runtime, the runtimes begin increasing very quickly with a learned sample fraction greater than 50\%. This is caused by a high density of samples being in small regions, leading to an increased number of nearest neighbors for each sample.
As the 50\% distribution performed well in all three factors, we use this as our default for this paper, and we observed similar results in other planning environments.

In this work we only consider sampling fractions that are constant over the duration of the planning algorithm. However, the learned sampling distribution typically result in rapid convergence (a few hundred samples) in most cases, with a small fraction of problems taking longer. In these cases, the learned samples fail to produce a trajectory, and gaps are filled via the auxiliary sampler. As a result, performance could likely be improved by first sampling primarily from the learned distribution, and increasingly sampling from the uniform distribution as the problem progresses. This is a relatively minor consideration, but may improve performance, especially when the amount of training data is limited.

\subsection{Learning Dependent Sample Sets}\label{sec:multiple} 

\begin{figure}[htb]
\centering
    \begin{subfigure}{0.32\textwidth}
        \includegraphics[width=\textwidth]{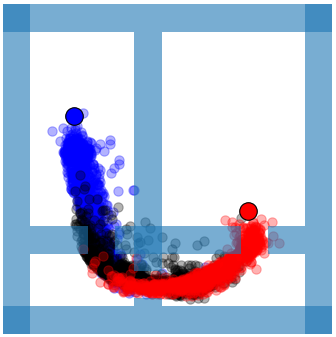}
        \caption{}
        \label{fig:multi1}
    \end{subfigure} 
    \begin{subfigure}{0.32\textwidth}
        \includegraphics[width=\textwidth]{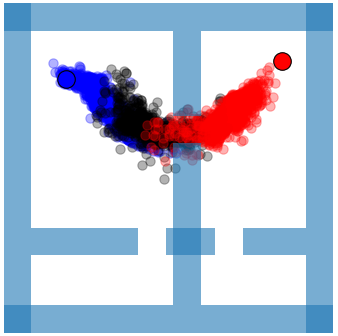}
        \caption{}
        \label{fig:multi2}
    \end{subfigure} 
    \begin{subfigure}{0.32\textwidth}
        \includegraphics[width=\textwidth]{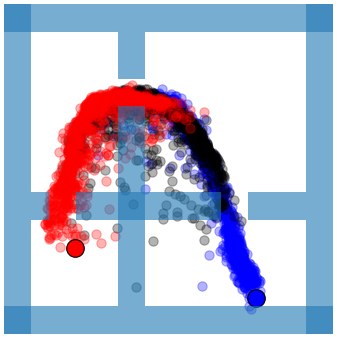}
        \caption{}
        \label{fig:multi3}
    \end{subfigure} 
    \begin{subfigure}{0.8\textwidth}
        \includegraphics[width=\textwidth]{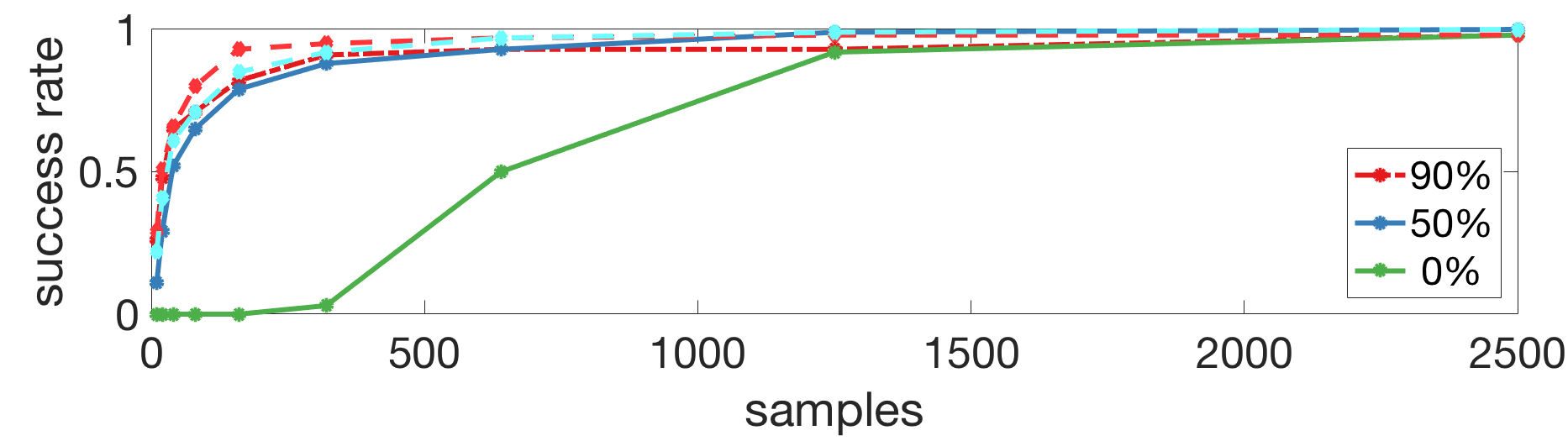}
        \caption{Convergence}
        \label{fig:narrowConvergenceMultiConvergence}
    \end{subfigure} 
\caption{(\ref{fig:multi1}-\ref{fig:multi3}) Learned distributions of multiple, dependent samples drawn together (each one in red, black, and blue), effectively enforcing some dispersion between them. (\ref{fig:narrowConvergenceMultiConvergence}) Success rates for single sample distributions and multi-sample distributions (lighter colors denote the multi-sample counterparts).
}
\label{fig:narrowPassageResultsMulti}
\end{figure}

To showcase the generality of the learned sampling distributions methodology and its ability to capture arbitrary and complex distributions, we use the proposed methodology to learn sets of samples -- meaning we learn a distribution of multiple samples at once, to be drawn in batches.
In this case, we learn from solution trajectories with three or more samples.
We are thus learning not only a distribution to model the promising regions of the state space, but multiple distributions at once with dependency between them (i.e., the methodology learns to disperse the samples along the trajectory).
Fig. \ref{fig:narrowPassageResultsMulti} shows resulting distributions and the success rate of this method compared to learning only a single sample. 
As expected, the distributions learn to be well-distributed along the solution trajectory, resulting in higher success rates (e.g., the success rate for a 90\% ratio of learned samples to uniform samples increased from 82\% to 93\% at 100 samples).
This result corroborates the findings of \citet{janson2018deterministic}, which found the primary benefit of low-dispersion sampling is in finding solutions with fewer samples.

\subsection{Varied Obstacle Density}

In the previous section the learned sampling distributions methodology was shown to generalize well to previously unseen problem instances (new initial states, goal regions, and obstacle sets).
In these cases, the test and train problem sets are drawn from the same problem generator.
This section investigates the performance of this method when the test problems are significantly different from those seen during the training phase.
The ability for machine learning systems to generalize (or extrapolate beyond training data) is a current active topic of research. 
While we anticipate future developments will enable better generalization, in this subsection we aim to characterize the out-of-distribution generalization of our proposed methodology.

\begin{figure}[htb]
\centering
    \begin{subfigure}{0.25\textwidth}
        \includegraphics[width=\textwidth]{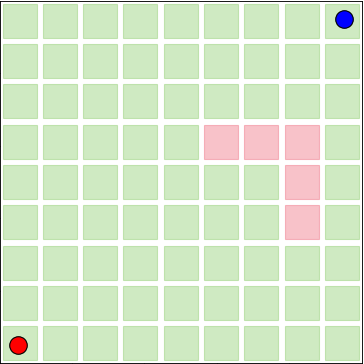}
        \caption{Low Complexity}
        \label{fig:mazeComplex0}
    \end{subfigure}
    \begin{subfigure}{0.25\textwidth}
        \includegraphics[width=\textwidth]{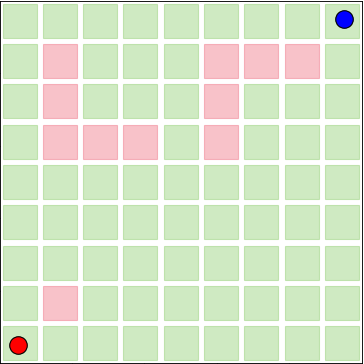}
        \caption{Medium Complexity}
        \label{fig:mazeComplex4}
    \end{subfigure}
    \begin{subfigure}{0.25\textwidth}
        \includegraphics[width=\textwidth]{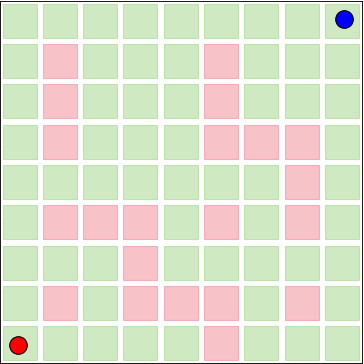}
        \caption{High Complexity}
        \label{fig:mazeComplex8}
    \end{subfigure}
\caption{Representative mazes for each complexity with initial states and goal regions held constant.
}
\label{fig:mazeComplex}
\end{figure}

\begin{figure}[ht]
\centering
    \begin{subfigure}{0.48\textwidth}
        \includegraphics[width=\textwidth]{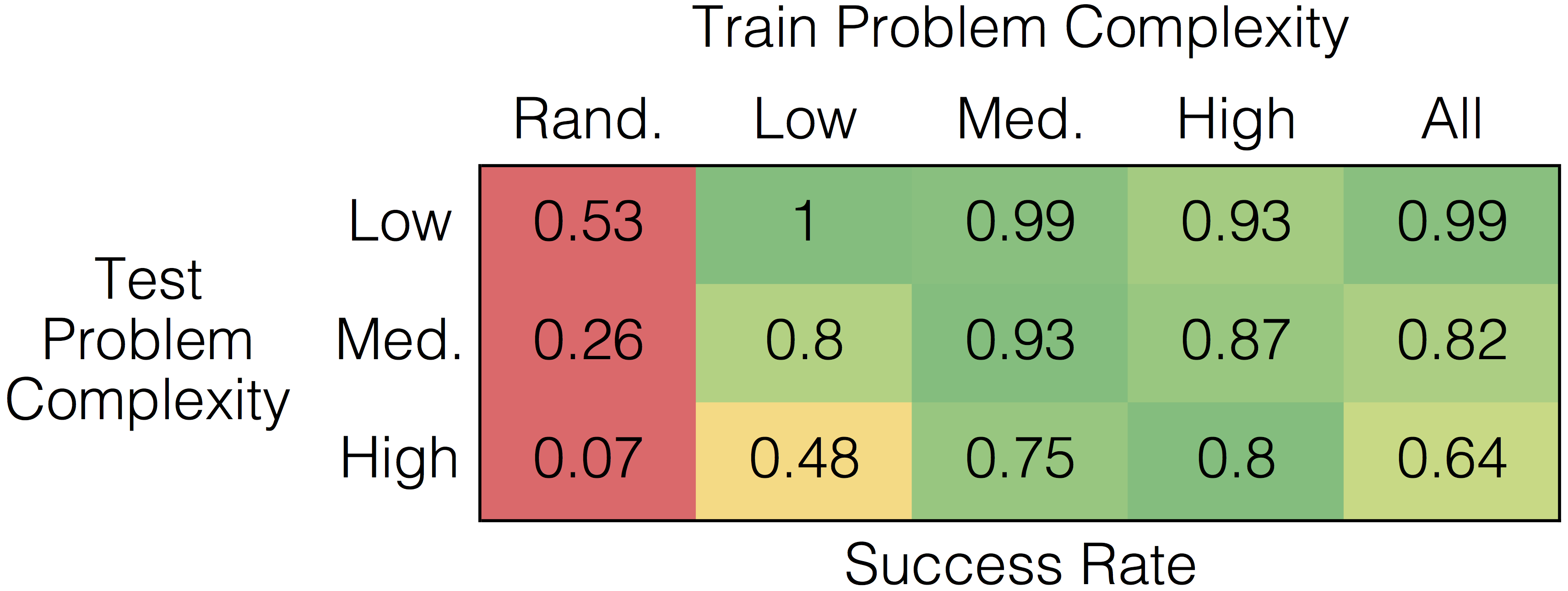}
        \caption{Success Rate at 500 Samples}
        \label{fig:mazeGeneralizationSuccess}
    \end{subfigure}
    \begin{subfigure}{0.48\textwidth}
        \includegraphics[width=\textwidth]{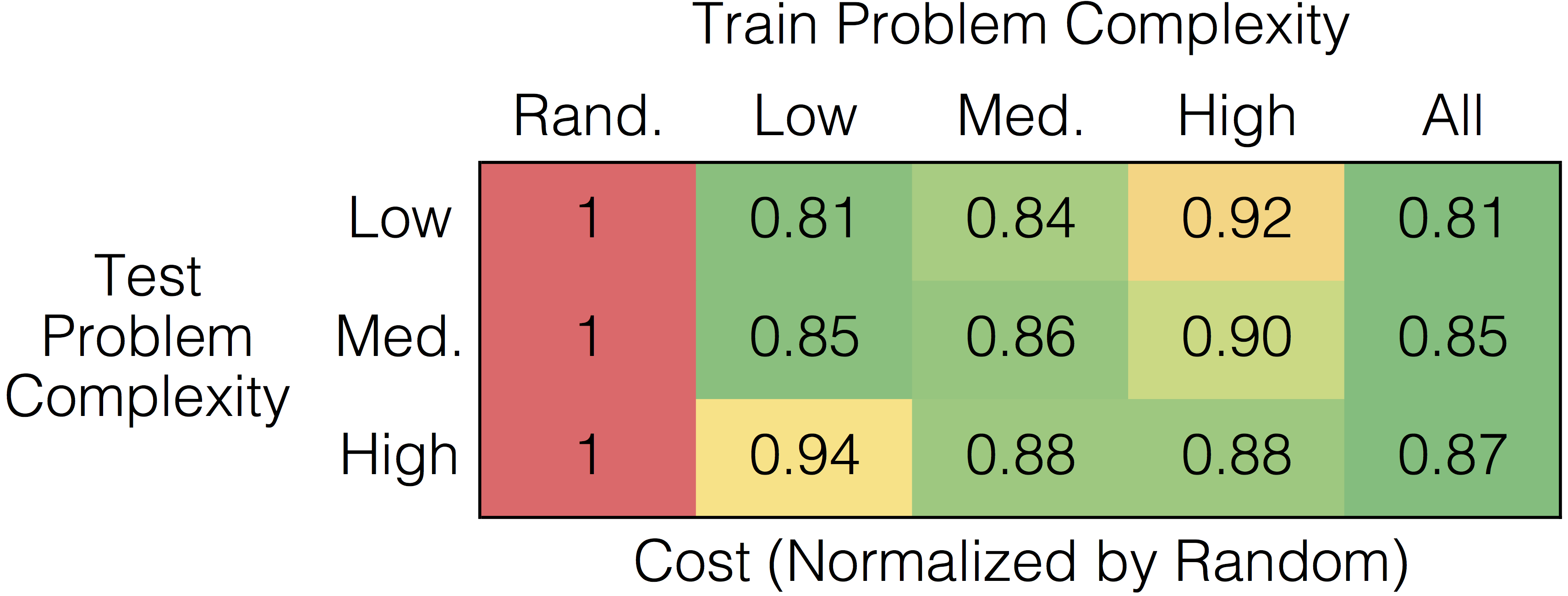}
        \caption{Normalized Cost at 4000 Samples}
        \label{fig:mazeGeneralizationCost}
    \end{subfigure}
\caption{Results for generalization across testing environments. For both grids, the columns denote the complexity of the training problems, where \textit{All} denotes problems drawn with equal probability from each class, and \textit{Rand.} denotes using a uniform sampling distribution. The rows denote the performance on low, medium, and high complexity planning problems. All training complexities outperformed random sampling on all test complexities, both in terms of success rate and normalized cost. Training on the same complexity level as used in testing always achieves the best (or nearly the best, in the case of cost) performance. Training on low complexity problems and testing on high complexity problems is potentially problematic, as the optimal trajectories are close to the shortest path in free space, and thus the CVAE model does not learn to effectively condition on the obstacles. These results show that it is important in practice to ensure that the training data used is reasonably representative of the test conditions.
}
\label{fig:mazeGeneralization}
\end{figure}

Our approach is as follows: we generate maze-like environments randomly, from a random maze generator (described below) that takes maze complexity as an argument. 
We generate training data for three different complexity levels (low, medium, and high), and train sampling distributions on these datasets. 
Then, we investigate the performance of double integrator systems (conditioned on an occupancy grid of the environment) on planning problems of some complexity, with sampling distributions trained from a dataset of a different complexity. Concretely, we train a sampling distribution on each of the low, medium, and high complexity datasets, and test on each of them as well. 
We do not test on the train dataset, different test and train datasets are generated for each complexity level. 
In addition to this, we also compare against a uniform sampling distribution and a CVAE trained on all three complexity levels. 
Examples of each complexity level are plotted in Fig. \ref{fig:mazeComplex}. Results are plotted in Fig. \ref{fig:mazeGeneralization}. 
In our experiments, we found that for all cases the learned sampling distribution substantially outperformed a standard uniform distribution.
Of the learned distributions, we found that the worst performance was achieved when distributions were trained on low complexity environments and tested on high complexity environments. This is fairly intuitive, as low complexity environments have few obstacles, and the distribution is heavily biased toward samples in the center of the workspace with velocities toward the goal. The best performance was (roughly) achieved when the train and test complexity were the same. 

\paragraph{Maze Generation. } The maze generation algorithm was implemented on a square grid with an odd number of rows and columns. The complexity of the maze generating function was indexed by two numbers: the number of obstacles generated ($n$) and the number of steps taken ($m$). Referring to these with tuples $(n,m)$, low, medium, and high complexity corresponded to $(2,2)$, $(4,2)$, and $(6,4)$. The maze was generated by sampling points in the grid on odd-indexed cells, sampling a random direction, and attempting to take a step, where each step corresponded to two cells. The cells between the previous point and the new step would then be added to the obstacle. This continued while less than $m$ steps were taken for that obstacle. If the cell was occupied, then the step was not taken. This was repeated for a total of $n$ obstacles. If the initial sampled point of an obstacle was in another obstacle, it was not resampled. This process was applied to an $11 \times 11$ grid, and the outer region layer of one cell was discarded. This was performed because this maze generation process often left this space without obstacles, so motion planning algorithms could find simple feasible paths by sampling along the edge of the grid. 

\subsection{Iterative Training of CVAE}

\begin{figure}[htb]
\centering
    \includegraphics[width=0.8\textwidth]{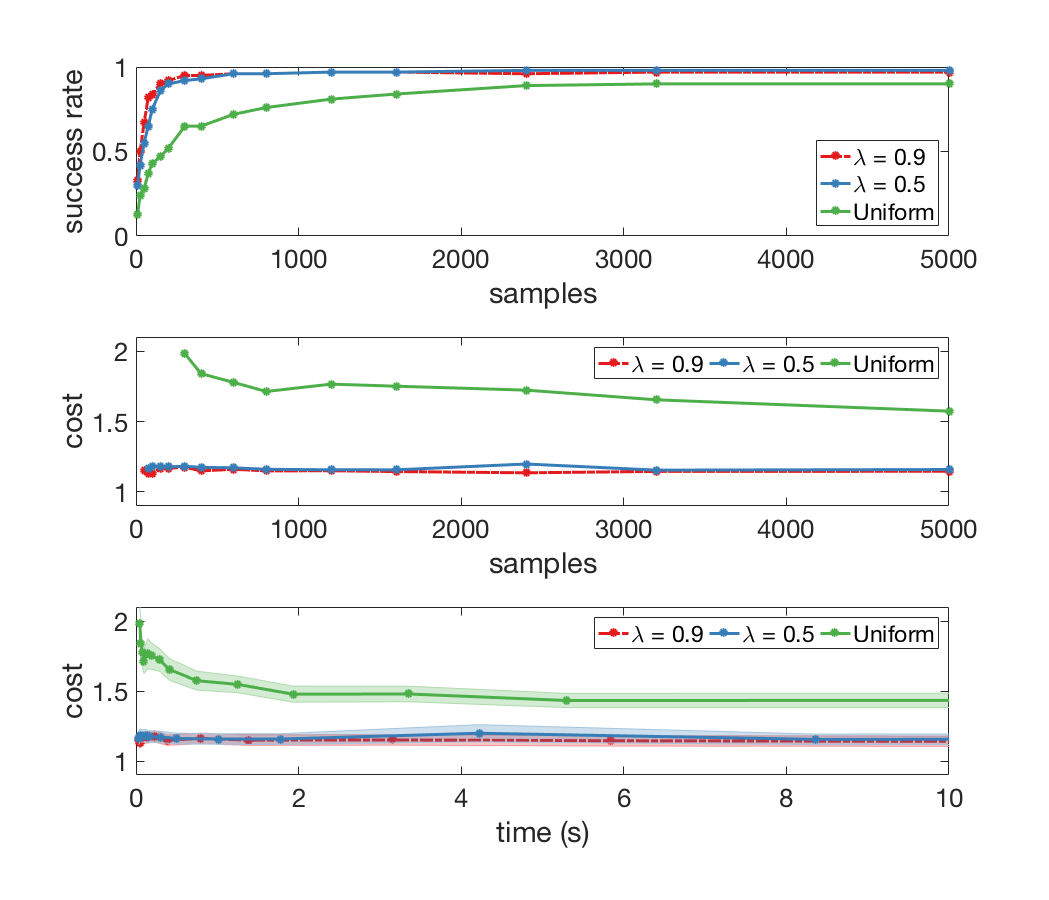}
    \caption{Performance of the learned sampling framework on the multirobot motion planning problem.
    }
\label{fig:multirobotConvergence}
\end{figure}

Fig. \ref{fig:multirobotConvergence} shows the performance of the approach on a multirobot motion planning problem. 
This problem consists of planning for three single integrator robotic systems in their joint state space. 
Empirically, we found learning near-optimal sampling distributions from successful motion plans was challenging for this problem because the motion plans generated through uniform sampling were of low quality (a high quality plan for the problem requires not only samples in the correct workspace locations, but synchronized along the trajectory of each robot).
The difficulty of generating high quality trajectories for this system using uniform sampling distributions is shown in Fig. \ref{fig:multirobotConvergence}, where after 5000 samples the best generated trajectory has a normalized cost of more than 1.5. 

\begin{figure}[htb]
\centering
    \begin{subfigure}{0.3\textwidth}
        \includegraphics[width=\textwidth]{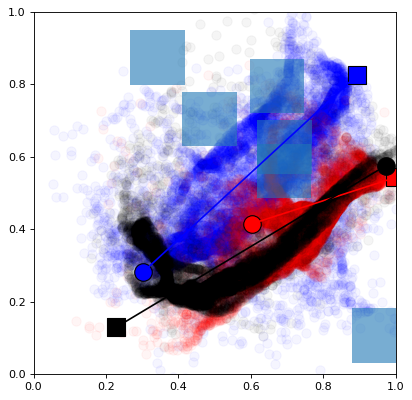}
        \caption{Trained on data from uniform sampling}
        \label{fig:multirobotDist_V0}
    \end{subfigure}
    \begin{subfigure}{0.3\textwidth}
        \includegraphics[width=\textwidth]{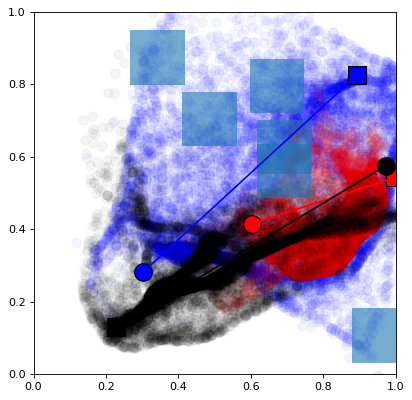}
        \caption{Trained on data from Fig. \ref{fig:multirobotDist_V0}}
        \label{fig:multirobotDist_V2}
    \end{subfigure}
    \begin{subfigure}{0.3\textwidth}
        \includegraphics[width=\textwidth]{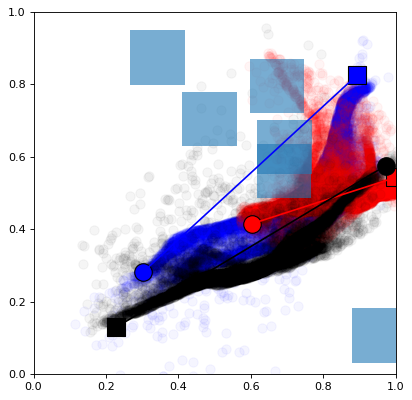}
        \caption{Trained on data from Fig. \ref{fig:multirobotDist_V2}}
        \label{fig:multirobotDist_V3}
    \end{subfigure}
\caption{ Learned sampling distributions for a varying number of phases of re-generating training datasets using the learned sampling distribution. 
The different colors correspond to different robots, and the planning problem is in their joint state space. 
The straight lines plotted in each figure connect the start state to the goal state for each robot. 
Note that iteratively regenerating training data allows rapid convergence of the learned sampling distributions.
}
\label{fig:multirobotDist}
\end{figure}

A na\"ive approach to this problem would be to simply generate a very large amount of data, and increase training time. However, data gathering efficiency can be dramatically improved by iteratively updating the sampling scheme used, as opposed to solely using the uniform sampling distribution for gathering training data. Figs. \ref{fig:multirobotDist_V2}, \ref{fig:multirobotDist_V3} show the sampling distributions after retraining with data generated from a previous (less-optimal) learned sampling distribution. We found that this dramatically improved sample complexity. 

\subsection{Human Demonstration}

\begin{figure}[htb]
\centering
    \begin{subfigure}{0.3\textwidth}
        \includegraphics[width=\textwidth]{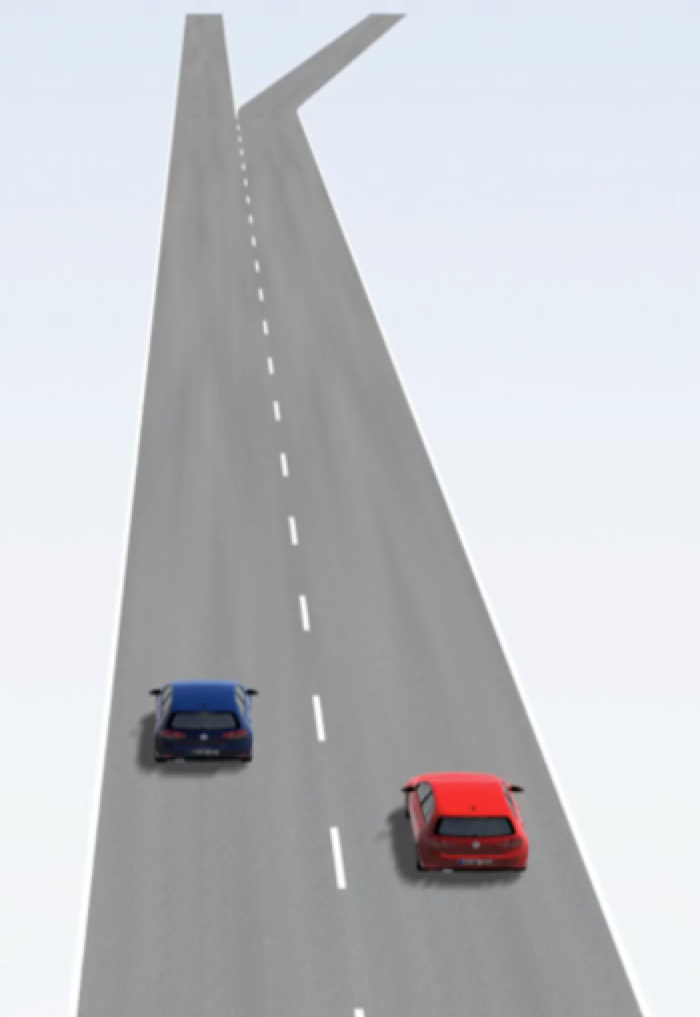}
        \caption{}
        \label{fig:carReal}
    \end{subfigure}
    \begin{subfigure}{0.15\textwidth}
        \includegraphics[width=\textwidth]{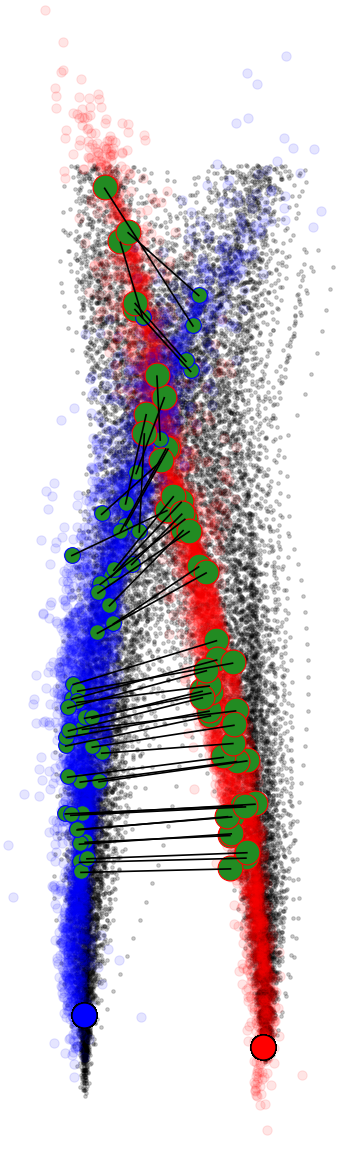}
        \caption{}
        \label{fig:carXY}
    \end{subfigure}
    \begin{subfigure}{0.15\textwidth}
        \includegraphics[width=\textwidth]{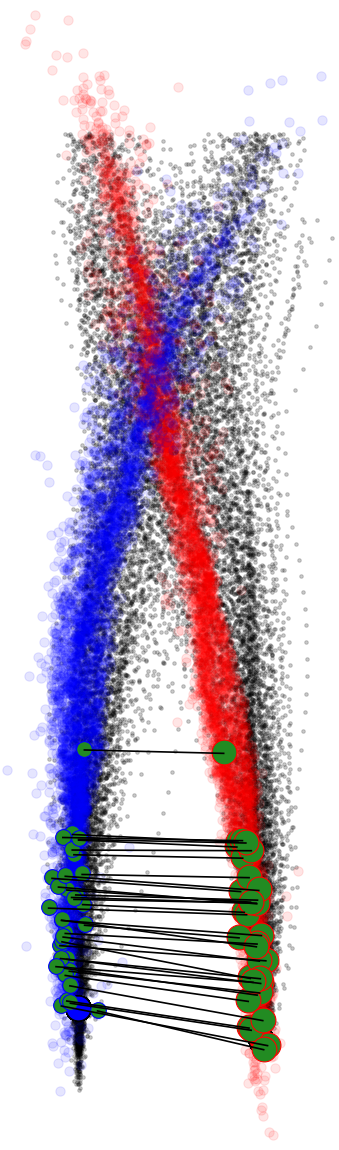}
        \caption{}
        \label{fig:carXYtail}
    \end{subfigure}
\caption{(\ref{fig:carReal}) Setup for lane change problem, generated from human demonstration, where the red and blue car must switch lanes and make their respective exits. (\ref{fig:carXY}-\ref{fig:carXYtail}) The training dataset is visualized in black, overlayed with learned distributions for the blue and red initial car states. Displayed are the $x$ and $y$ positions of each car at several samples. Each line connects the position of the two cars at a given time.
Initially, the red car is behind the blue car, but at a higher velocity and thus eventually passes the blue car in most samples (the red car is leading samples visualized in (\ref{fig:carXY}) and tailing in (\ref{fig:carXYtail})).
Note each sample is not in self collision.}
\label{fig:car}
\end{figure}

Finally, we demonstrate the methodology on a problem trained from a data source other than computed motion plans -- human demonstration. 
The problem is a multi-robot planning problem consisting of two cars completing a lane changing maneuver (Fig. \ref{fig:carReal}), with a total of 22 states, as well as a constraint on the two cars colliding and a preference towards the cars remaining in their lanes when not changing lanes. 
The data was collected from human demonstration on a two person driving simulator \citep{schmerling2017multimodal}. 
Because this problem is beyond the reach of sampling-based motion planning with uniform samples and because we do not have access to a two point boundary value problem solver, we only examine this distribution qualitatively.
The resulting learned distribution is visualized for a single initial state in Fig. \ref{fig:carXY} and overlayed on the total dataset. 
The learned distribution effectively encapsulates the necessary factors for a successful motion plan: the collision constraint, the preference towards the center of lanes, the preference to maintain forward velocity, and the choice of states applicable to a lane change maneuver.

\section{Discussion and Conclusions}\label{sec:conclusions}

\paragraph{Conclusions.} In this paper we have presented a methodology to bias samples in the state space for sampling-based motion planning algorithms. 
In particular, we have used a conditional variational autoencoder to learn subspaces of valid or desirable states, conditioned on problem details such as obstacles. 
We have compared our methodology to several state of the art methods for sample biasing, and have demonstrated it on multiple systems, showing approximately an order of magnitude improvement in cost and success rate over uniform sampling. 
This learning-based approach is promising due to its generality and extensibility, as it can be applied to any system and can leverage any problem information available. 
Its ability to automatically discover useful representations for motion planning (as seen by the near immediate convergence) is similar to recent results from deep learning in the computer vision community, which require less human intuition and handcrafting, and exhibit superior performance.

\paragraph{Future Work.} There are many possible avenues open for future research. 
One promising extension is the incorporation of semantic workspace information through the conditioning variable. 
These semantic maps show promise towards allowing mobile robots to better understand task specifications and interact with humans \citep{kostavelis2015semantic}.
Another promising extension builds upon recent work demonstrating the favorable theoretical properties and improved performance of non-independent samples \citep{janson2018deterministic}. 
An approach to reducing the independence of the samples was presented in Section \ref{sec:workspace}, but extensions beyond this exist, including generating large sample sets (e.g., $>1000$).
Lastly, for systems with constraints that force valid configurations to lie on zero-measure manifolds, recent work has focused on projective methods \citep{jaillet2013path}. 
Our methodology, while not capable of sampling directly on this manifold, can easily learn to sample near it, and therefore, potentially dramatically improve the performance of these methods. 

In this work we have explored conditioning on workspace maps. However, this was performed only for relatively small, planar problems. Scaling this to large problems is challenging, as the number of conditioning variables grows exponentially with the occupancy grid resolution and with the problem dimensionality. However, our experiments provided promising evidence that this conditioning could be used to substantially improve performance. As such, a promising future line of work is investigating more efficient methods for environment conditioning. One possible approach is finding environment representations that are relatively low dimensional, and encode useful information for motion planning. Another is to instead switch to an adaptive scheme. 
In this approach, instead of conditioning on the workspace map directly, one could condition on recent samples with known collision test results or one could condition on the state of the tree itself. 
This gives a constantly updated set of conditioning variables corresponding to the structure of the workspace. However, this method also has associated scalability questions that merit future work. 

\paragraph{Application in Practice.}
Note again, the goal of this work is to compute a distribution representing promising regions (i.e., regions where optimal motion plans are likely to be found) through a learned latent representation of the system conditioned on the planning problem.
During the offline CVAE training phase, we recommend training from optimal motion plans to best demonstrate promising regions.
We generally train on the order of one hundred thousand motion plans, the generation of which can be accelerated with approximately optimal, GPU-based planning algorithms \citep{ichter2017group}.
To form the conditional variable, we recommend including all available problem information, particularly the initial state and goal region. 
If available, workspace obstacles can be easily included through occupancy grids; we note however that obstacles can be included in any form, as the neural network representation for the CVAE has the potential to arbitrarily project these obstacles as necessary.
As mentioned in Section \ref{sec:method}, we use a weighting term on the KL divergence in the ELBO for training, as in e.g. \citet{higgins2017beta}. 
This was chosen independently for each planning problem, in the range of $10^{-4}$ to $10^{-2}$. 
This choice was simply based on visual inspection of the samples generated by the trained model -- in practice, a more thorough evaluation could be performed via evaluating the effect of the KL weighting on the performance of the planning algorithm. 
Multiple {\em dependent} samples can also be generated at once (Section \ref{sec:multiple}); we found even as few as three resulted in marked increases in success rate.
Finally, in the online phase of the algorithm, we draw a combination of learned and uniform samples to ensure good state space coverage; we found a 50-50 split to be most effective.

\section*{Acknowledgment}
 This work was supported by a Qualcomm Innovation Fellowship and by NASA under the Space Technology Research Grants Program, Grant NNX12AQ43G.
  James Harrison was supported in part by the Stanford Graduate Fellowship and the National Sciences and Engineering Research Council (NSERC).
The GPUs used for this research were donated by the NVIDIA Corporation.
The authors also wish to thank Edward Schmerling for helpful discussions.


\bibliographystyle{agsm} 
\bibliography{cite} 

\end{document}